\documentclass[sn-mathphys,Numbered]{sn-jnl}% Math and Physical Sciences Reference Style
\usepackage{graphicx}%
\usepackage{multirow}%
\usepackage{amsmath,amssymb,amsfonts}%
\usepackage{amsthm}%
\usepackage{mathrsfs}%
\usepackage[title]{appendix}%
\usepackage{xcolor}%
\usepackage{textcomp}%
\usepackage{manyfoot}%
\usepackage{booktabs}%
\usepackage{algorithm}%
\usepackage{algorithmic}%
\usepackage{listings}%
\usepackage{subcaption}%
\usepackage{bm}%
\theoremstyle{thmstyleone}%
\theoremstyle{thmstyletwo}%

\theoremstyle{thmstylethree}%
\newtheorem{definition}{Definition}%

\begin{document}

\title[Geometry-Complete Perceptron Networks for 3D Molecular Graphs]{Geometry-Complete Perceptron Networks for \\ 3D Molecular Graphs}


\author*[1]{\fnm{Alex} \sur{Morehead}}\email{acmwhb@missouri.edu}
\author*[1]{\fnm{Jianlin} \sur{Cheng}}\email{chengji@missouri.edu}

\affil*[1]{\orgdiv{Electrical Engineering \& Computer Science}, \orgname{University of Missouri-Columbia}, \orgaddress{\street{W1024 Lafferre Hall}, \city{Columbia}, \postcode{65211}, \state{Missouri}, \country{United States of America}}}


\abstract{The field of geometric deep learning has had a profound impact on the development of innovative and powerful graph neural network architectures. Disciplines such as computer vision and computational biology have benefited significantly from such methodological advances, which has led to breakthroughs in scientific domains such as protein structure prediction and design. In this work, we introduce \textsc{GCPNet}, a new geometry-complete, SE(3)-equivariant graph neural network designed for 3D molecular graph representation learning. Rigorous experiments across four distinct geometric tasks demonstrate that \textsc{GCPNet}'s predictions (1) for protein-ligand binding affinity achieve a statistically significant correlation of 0.608, more than 5\% greater than current state-of-the-art methods; (2) for protein structure ranking achieve statistically significant target-local and dataset-global correlations of 0.616 and 0.871, respectively; (3) for Newtownian many-body systems modeling achieve a task-averaged mean squared error less than 0.01, more than 15\% better than current methods; and (4) for molecular chirality recognition achieve a state-of-the-art prediction accuracy of 98.7\%, better than any other machine learning method to date. The source code, data, and instructions to train new models or reproduce our results are freely available at \href{https://github.com/BioinfoMachineLearning/GCPNet}{https://github.com/BioinfoMachineLearning/GCPNet}.}




\maketitle

\section{Introduction}\label{section:introduction}

\begin{figure}[t]
    \centering
    \begin{subfigure}{0.31\columnwidth}
        \centering
        \textbf{(a)}
        \includegraphics[width=\columnwidth]{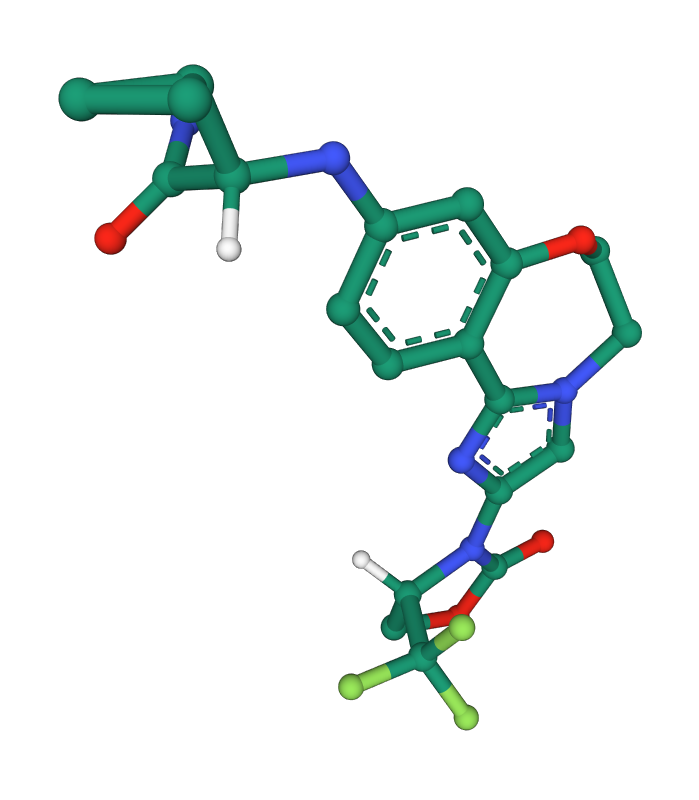}
        \caption*{PDB ID: 8EXU}
    \end{subfigure}
    \begin{subfigure}{0.31\columnwidth}
        \centering
        \textbf{(b)}
        \vspace{0.3cm}
        \includegraphics[width=\columnwidth]{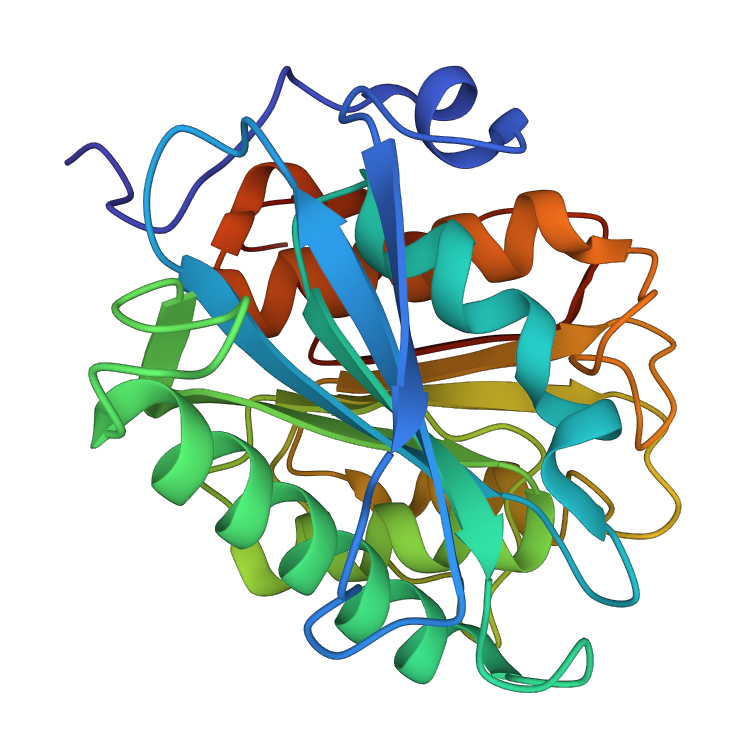}
        \caption*{PDB ID: 8GU4}
    \end{subfigure}
    \begin{subfigure}{0.31\columnwidth}
        \centering
        \textbf{(c)}
        \vspace{0.55cm}
        \includegraphics[width=\columnwidth]{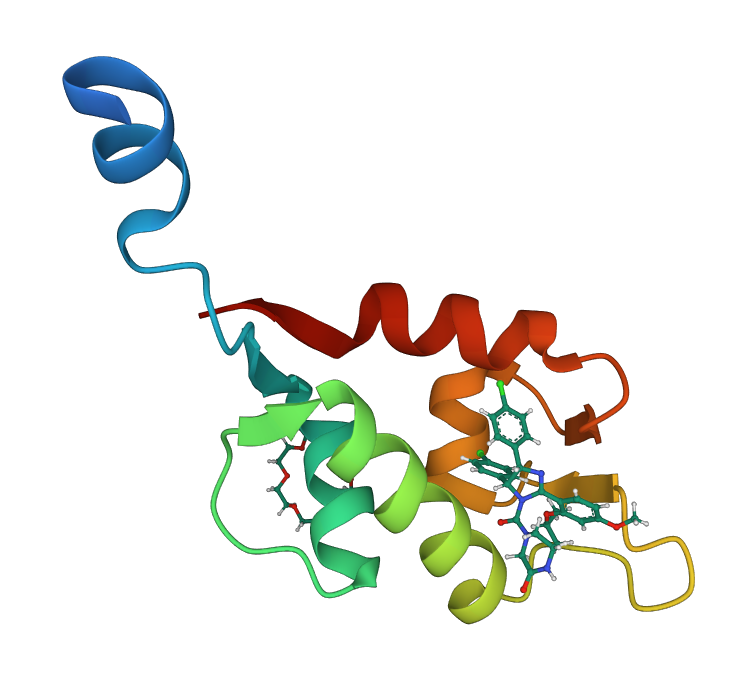}
        \caption*{PDB ID: 8HDG}
    \end{subfigure}
    
    \caption{Mol* \cite{sehnal2021mol} visualizations of various kinds of representative 3D molecular structures studied in this work including \textbf{(a)} small molecules (i.e., ligands), \textbf{(b)} proteins, and \textbf{(c)} protein-ligand complexes.}
    \label{figure:3d_molecules}
\end{figure}

Over the last several years, the field of deep learning has pioneered many new methods designed to process graph-structured inputs. Being a ubiquitous form of information, graph-structured data arises from numerous sources such as the fields of physics and chemistry, as shown in Figure \ref{figure:3d_molecules}. Moreover, the relational nature of graph-structured data allows one to identify and characterize topological associations between entities in large real-world networks (e.g., social networks).

In particular, 3D data often emerges in domains such as computer vision and can be readily described as graph-structured inputs \cite{valsesia2018learning}. Studies such as those of \cite{qi20173d, zhang2018graph, zhou2021adaptive} have demonstrated the utility of this approach to modeling 3D data as graphs. Additionally, to process and analyze such 3D information in a meaningful, powerful, and concise way, one must also carefully consider the symmetries present in such data to reduce the geometric redundancies they might present to a machine learning model \cite{esteves2020theoretical}.

Many disciplines present types of data for which 3D geometric information can be carefully analyzed to produce meaningful and reliable predictions about the system at hand. For example, in protein biology, knowing the 3D structure of a protein macromolecule is a key step towards developing a deeper understanding of its molecular function in living organisms \cite{hegyi1999relationship}. In a context-specific manner, similar geometric insights have been proposed in fields such as neurobiology \cite{umulis2012importance} and materials design \cite{matter2022importance}. In light of such insights, the field of deep learning has grown to account for the importance of geometry in the representation learning of real-world objects.

In the case of being presented with geometric (non-Euclidean domain) data, machine learning systems have been developed to process arbitrarily-structured inputs such as graphs and meshes \cite{hamilton2020graph, cao2022geometric}. A subfield of geometric machine learning, geometric deep learning, has recently received much attention from researchers for its various success cases in using deep neural networks to faithfully model geometric data in the form of 3D graphs and manifolds \cite{masci2016geometric, bronstein2017geometric}.

Previous works in geometric deep learning \cite{bronstein2021geometric} have explored the use of neural networks for modeling physical systems \cite{cao2020comprehensive, kiarashinejad2020knowledge, james2021citywide}. Some of the earliest neural networks applied to physical systems include convolutional networks (CNNs) \cite{lecun1995convolutional, he2016deep, he2017mask, krizhevsky2017imagenet}, graph neural networks (GNNs) \cite{kipf2016semi, gilmer2017neural, velivckovic2017graph, dwivedi2020generalization}, and point cloud neural networks \cite{qi2017pointnet, liu2019relation, zhang2019shellnet}. Likewise, recurrent neural networks (RNNs) have been used to classify sequential real-world data such as speech activity \cite{graves2013speech, graves2014towards} and have been adopted as case studies for how to develop next-generation wave-based electronics hardware \cite{hughes2019wave}.

In scientific domains such as computational biology and chemistry, graphs are often used to represent the 3D structures of molecules \cite{NIPS2015_f9be311e, liu2021spherical}, chemical compounds \cite{akutsu2013comparison}, and even large biomolecules such as proteins \cite{xia2021geometric, morehead2022egr}. Graphs have even been used in fields such as computational physics to model complex particle physics simulations \cite{shlomi2020graph} as well as in real-world traffic systems to predict travel times and delays \cite{derrow2021eta}. Underlying many of these successful examples of graph representations are GNNs, a class of machine learning algorithms specialized in processing irregularly-structured input data such as graphs. Careful applications of graph neural networks in scientific domains have considered the physical symmetries present in many scientific data such as molecular state symmetries \cite{ye2020symmetrical} or physical dynamics constraints \cite{han2022learning} and have leveraged such symmetries to design new attention-based neural network architectures \cite{morehead2022geometric, jumper2021highly}.

Throughout their development, geometric deep learning methods have expanded to incorporate within them equivariance to various geometric symmetry groups to enhance their generalization capabilities and adversarial robustness. Methods such as group-equivariant CNNs \cite{cohen2016group}, Tensor Field Networks \cite{Thomas2018TensorFN}, SE(3)-Transformers \cite{fuchs2020se}, and equivariant GNNs \cite{fuchs2020se, jing2020learning, jing2021equivariant, kofinas2021roto, gasteiger2021gemnet, schutt2021equivariant, huang2022equivariant, tholke2022equivariant, pmlr-v162-du22e, 10.1145/3534678.3539441, batzner20223} have paved the way for the development of future deep learning models that respect physical symmetries present in 3D data (e.g., rotation equivariance with respect to input data symmetries). Concurrently to these efforts, self-supervised learning methods have begun to facilitate automatic detection and enforcement of the symmetries present in input data within the network's representations for such inputs \cite{dangovski2021equivariant}. Nonetheless, deciding how to optimize self-supervised learning algorithms for one's desired level of equivariance has proven to be a challenging task \cite{xie2022should}.

In total, in this work, we make connections between geometric graph neural networks, equivariance, and geometric information-completeness guarantees that provide one with a rich foundation on which to build new graph neural network architectures. In particular, we introduce a new graph neural network model, \textsc{GCPNet}, that is equivariant to the group of 3D rotations and translations (i.e., SE(3), the special Euclidean group) and guarantees geometric information completeness following graph message-passing on 3D point clouds. We demonstrate its expressiveness and flexibility for modeling physical systems through rigorous experiments for distinct molecular-geometric tasks. In detail, we provide the following contributions:

\begin{itemize}
    \item In contrast to prior geometric networks for molecules that are insensitive to their chemical chirality, cannot detect global physical forces acting upon each atom, or do not directly learn geometric features, we present the first geometric graph neural network architecture with the following desirable properties for learning from 3D molecules: (1) the ability to directly predict translation and rotation-invariant scalar properties and rotation-equivariant vector-valued quantities for nodes and edges, respectively; (2) a rotation and translation-equivariant method for iteratively updating node positions in 3D space; (3) sensitivity to molecular chirality; and (4) a means by which to learn from and account for the global forces acting upon the atoms within its inputs.
    \item We establish new state-of-the-art results for four distinct molecular-geometric representation learning tasks - molecular chirality recognition, protein-ligand binding affinity prediction, protein structure ranking, and Newtonian many-body-systems modeling - where model predictions vary from analyzing individual nodes to summarizing entire graph inputs. \textsc{GCPNet}'s performance for these tasks is statistically significant and surpasses that of previous state-of-the-art machine learning methods for 3D molecules.
\end{itemize}

\section{Results}\label{section:results}

In this work, we consider four distinct modeling tasks comprised of seven datasets in total, where implementation details are discussed in Appendix \ref{section:implementation_details}. We note that additional experiments are included in Appendix \ref{section:additional_experiments_and_results} for interested readers.

\textbf{Assessing model sensitivity to molecular chirality.} Molecular chirality is an essential geometric property of 3D molecules for models to consider when making predictions for downstream tasks. Simply put, this property describes the "handedness" of 3D molecules, in that, certain molecules cannot be geometrically superimposed upon a mirror reflection of themselves using only 3D rotation and translation operations. This subsequently poses a key challenge for machine learning models: Can such predictive models effectively sensitize their predictions to the effects of molecular chirality such that, under 3D reflections, their molecular feature representations change accordingly? To answer this question using modern machine learning methods, we adopt the rectus/sinister (RS) 3D molecular dataset of \cite{adams2021learning} to evaluate the ability of state-of-the-art machine learning methods to distinguish between right-handed and left-handed versions of a 3D molecule. In addition, we carefully follow their experimental setup including dataset splitting and evaluation criteria, where we evaluate each method's classification accuracy in distinguishing between right and left-handed versions of a molecule. Baseline methods for this task include state-of-the-art invariant neural networks (INNs) and equivariant neural networks (ENNs), where we list each method's latest results for this task as reported in \cite{schneuing2023structurebased}.

\begin{table}[t]
    \caption{Comparison of \textsc{GCPNet} with baseline methods for the RS task. The results are averaged over three independent runs. The top-1 (best) results for this task are in \textbf{bold}, and the second-best results are \underline{underlined}.}
    \label{table:rs_results}
    \centering
    % \scriptsize  % Or footnotesize, scriptsize, tiny, etc.
    \begin{tabular}{lcc}
        & & \\ \midrule
        Type & Method & R/S Accuracy (\%) $\uparrow$ \\ \midrule
        INN & ChIRo \citeyearpar{schneuing2023structurebased} & \underline{98.5} \\
        & SchNet \citeyearpar{schneuing2023structurebased} & 54.4 \\
        & DimeNet++ \citeyearpar{schneuing2023structurebased} & 65.7 \\
        & SphereNet \citeyearpar{schneuing2023structurebased} & 98.2 \\ \midrule
        ENN & EGNN \citeyearpar{schneuing2023structurebased} & 50.4 \\
        & SEGNN \citeyearpar{schneuing2023structurebased} & 83.4 \\ \midrule
        Ours & \textsc{GCPNet} w/o Frames & 50.2 $\pm$ 0.6 \\
        & \textsc{GCPNet} & \textbf{98.7} $\pm$ \textbf{0.1} \\ \midrule
    \end{tabular}
\end{table}

\textbf{Contribution of frame embeddings for chirality sensitivity.} Table \ref{table:rs_results} shows that \textsc{GCPNet} is more accurately able to detect the effects of molecular chirality compared to all other baseline methods, even without performing any hyperparameter tuning. In particular, \textsc{GCPNet} outperforms ChIRo \cite{adams2021learning}, a GNN specifically designed to detect different forms of chirality in 3D molecules. Moreover, when we ablate \textsc{GCPNet}'s embeddings of local geometric frames, we find that this E(3)-equivariant (i.e., 3D rotation \textit{and} reflection-equivariant) version of \textsc{GCPNet} is no longer able to solve this important molecular recognition task, resulting in prediction accuracies at parity with random guessing. These two previous observations highlight that (1) \textsc{GCPNet}'s local frame embeddings are critical components of the model's sensitivity to molecular chirality and that, (2) using such frame embeddings, \textsc{GCPNet} can flexibly learn representations of 3D molecules that are more predictive of chemical chirality compared to hand-crafted methods for such tasks.

\begin{table}[t]
    \caption{Comparison of \textsc{GCPNet} with baseline methods for the LBA task. The results are averaged over three independent runs. The top-1 (best) results for this task are in \textbf{bold}, and the second-best results are \underline{underlined}.}
    \label{table:lba_results}
    \centering
    % \scriptsize  % Or footnotesize, scriptsize, tiny, etc.
    \begin{tabular}{lcccc}
        & & & & \\ \midrule
        Type & Method & RMSE \ $\downarrow$ & $p$\ $\uparrow$ & $Sp$\ $\uparrow$ \\ \midrule
        CNN & 3DCNN \citeyearpar{townshend2020atom3d} & 1.416 $\pm$ 0.021 & 0.550 & 0.553 \\
        & DeepDTA \citeyearpar{ozturk2018deepdta} & 1.866 $\pm$ 0.080 & 0.472 & 0.471 \\
        & DeepAffinity \citeyearpar{karimi2019deepaffinity} & 1.893 $\pm$ 0.650 & 0.415 & 0.426 \\ \midrule
        RNN & Bepler and Berger \citeyearpar{bepler2018learning} & 1.985 $\pm$ 0.006 & 0.165 & 0.152 \\
        & TAPE \citeyearpar{NEURIPS2019_37f65c06} & 1.890 $\pm$ 0.035 & 0.338 & 0.286 \\
        & ProtTrans \citeyearpar{elnaggar2021prottrans} & 1.544 $\pm$ 0.015 & 0.438 & 0.434 \\ \midrule
        GNN & GCN \citeyearpar{townshend2020atom3d} & 1.601 $\pm$ 0.048 & 0.545 & 0.533 \\
        & DGAT \citeyearpar{nguyen2021graphdta} & 1.719 $\pm$ 0.047 & 0.464 & 0.472 \\
        & DGIN \citeyearpar{nguyen2021graphdta} & 1.765 $\pm$ 0.076 & 0.426 & 0.432 \\
        & DGAT-GCN \citeyearpar{nguyen2021graphdta} & 1.550 $\pm$ 0.017 & 0.498 & 0.496 \\
        & MaSIF \citeyearpar{gainza2020deciphering} & 1.484 $\pm$ 0.018 & 0.467 & 0.455 \\
        & IEConv \citeyearpar{hermosilla2021intrinsicextrinsic} & 1.554 $\pm$ 0.016 & 0.414 & 0.428 \\
        & Holoprot-Full Surface \citeyearpar{somnath2021multiscale} & 1.464 $\pm$ 0.006 & 0.509 & 0.500 \\
        & Holoprot-Superpixel \citeyearpar{somnath2021multiscale} & 1.491 $\pm$ 0.004 & 0.491 & 0.482 \\
        & ProNet-Amino-Acid \citeyearpar{wang2023learning} & 1.455 $\pm$ 0.009 & 0.536 & 0.526 \\
        & ProNet-Backbone \citeyearpar{wang2023learning} & 1.458 $\pm$ 0.003 & 0.546 & 0.550 \\
        & ProNet-All-Atom \citeyearpar{wang2023learning} & 1.463 $\pm$ 0.001 & 0.551 & 0.551 \\
        & GeoSSL-DDM \citeyearpar{liu2023molecular} & 1.451 $\pm$ 0.030 & \underline{0.577} & \underline{0.572} \\ \midrule
        ENN & Cormorant \citeyearpar{anderson2019cormorant} & 1.568 $\pm$ 0.012 & 0.389 & 0.408 \\
        & PaiNN \citeyearpar{schutt2021equivariant} & 1.698 $\pm$ 0.050 & 0.366 & 0.358 \\
        & ET \citeyearpar{tholke2022equivariant} & 1.490 $\pm$ 0.019 & 0.564 & 0.532 \\
        & GVP \citeyearpar{jing2021equivariant} & 1.594 $\pm$ 0.073 & 0.434 & 0.432 \\
        & GBP \citeyearpar{10.1145/3534678.3539441} & \underline{1.405} $\pm$ \underline{0.009} & 0.561 & 0.557 \\ \midrule
        Ours & \textsc{GCPNet} w/o Frames & 1.485 $\pm$ 0.015 & 0.521 & 0.504 \\
        & \textsc{GCPNet} w/o \textsc{ResGCP} & 1.514 $\pm$ 0.008 & 0.471 & 0.468 \\
        & \textsc{GCPNet} w/o Scalars & 1.685 $\pm$ 0.000 & 0.050 & 0.000 \\
        & \textsc{GCPNet} w/o Vectors & 1.727 $\pm$ 0.005 & 0.270 & 0.304 \\
        & \textsc{GCPNet} & \textbf{1.352} $\pm$ \textbf{0.003} & \textbf{0.608} & \textbf{0.607} \\ \midrule
    \end{tabular}
\end{table}

\textbf{Evaluating predictions of protein-ligand binding affinity.} Protein-ligand binding affinity (LBA) prediction challenges methods to estimate the binding affinity of a protein-ligand complex as a single scalar value \cite{townshend2020atom3d}. Accurately estimating such values in a matter of seconds using a machine learning model can provide invaluable and timely information in the typical drug discovery pipeline \cite{rezaei2020deep}. The corresponding dataset for this SE(3)-invariant task is derived from the ATOM3D dataset \cite{townshend2020atom3d} and is comprised of 4,463 nonredundant protein-ligand complexes, where cross-validation splits are derived using a strict 30\% sequence identity cutoff. Results are reported in terms of the root mean squared error (RMSE), Pearson's correlation ($p$), and Spearman's correlation ($Sp$) between a method's predictions on the test dataset and the corresponding ground-truth binding affinity values represented as $pK = -\log_{10}(K)$, where $K$ is the binding affinity measured in Molar units. Baseline comparison methods for this task include a variety of state-of-the-art CNNs, recurrent neural networks (RNNs), GNNs, and ENNs.

The results shown in Table \ref{table:lba_results} reveal that, in operating on atom-level protein-ligand graph representations, \textsc{GCPNet} achieves the best performance for predicting protein-ligand binding affinity by a significant margin, notably improving performance across all metrics by 7\% on average. Here, to the best of our knowledge, \textsc{GCPNet} is also the first method capable of achieving Pearson and Spearman binding affinity correlations greater than 0.6 on the PDBBind dataset \cite{wang2005pdbbind} when employing a strict 30\% sequence identity cutoff. Moreover, we find that these correlations are highly statistically significant (i.e., Pearson's p-value of $2e-50$, Spearman's p-value of $2e-49$, and Kendall's tau correlation of 0.432 with a p-value of $3e-45$).

\textbf{Ablating network components reveals impact of model design.} Denoted as "\textsc{GCPNet} w/o ..." in Table \ref{table:lba_results}, our ablation studies with \textsc{GCPNet} for the LBA task demonstrate the contribution of each component in its model design. In particular, our proposed local frame embeddings improve \textsc{GCPNet}'s performance by more than 15\% across all metrics (\textsc{GCPNet} w/o Frames), where we hypothesize these performance improvements come from using these frame embeddings to enhance the model's sensitivity to molecular chirality. Similarly, our proposed residual $\textsc{GCP}$ module (i.e., \textsc{ResGCP}) improves \textsc{GCPNet}'s performance by 23\% on average.

Specifically of interest is the observation that independent removal of scalar and vector-valued features within \textsc{GCPNet} appears to severely decrease \textsc{GCPNet}'s performance for LBA prediction. Notably, removing the model's access to scalar-valued features degrades performance by 70\% on average, while not allowing the model to access vector-valued features reduces performance by 42\% on average. One possible explanation for these observations is that both types of feature representations the baseline \textsc{GCPNet} model learns (i.e., scalars and vectors) are useful for understanding protein-ligand interactions. In addition, our ablation results in Table \ref{table:lba_results} suggest that our proposed frame embeddings and \textsc{ResGCP} module are complementary to these scalar and vector-valued features in the context of predicting the binding affinity of a protein-ligand complex.

\begin{table}[t]
    \caption{Comparison of \textsc{GCPNet} with baseline methods for the PSR task. Local metrics are averaged across target-aggregated metrics. The best results for this task are in \textbf{bold}, and the second-best results are \underline{underlined}. N/A denotes a metric that could not be computed.}
    \label{table:psr_results}
    \centering
    % \scriptsize  % Or footnotesize, scriptsize, tiny, etc.
    \begin{tabular}{lcccccc}
        & & & & & & \\ \midrule
        & & Local & & & Global & \\
        \cmidrule(lr){2-4}\cmidrule(lr){5-7}
        Method & $p$\ $\uparrow$ & $Sp$\ $\uparrow$ & $K$ \ $\uparrow$ & $p$\ $\uparrow$ & $Sp$\ $\uparrow$ & $K$ \ $\uparrow$ \\ \midrule
        3DCNN \citeyearpar{townshend2020atom3d} & 0.557 & 0.431 & 0.308 & 0.780 & 0.789 & 0.592 \\
        GCN \citeyearpar{townshend2020atom3d} & 0.500 & 0.411 & 0.289 & 0.747 & 0.750 & 0.547 \\
        ProQ3D \citeyearpar{uziela2017proq3d} & 0.444 & 0.432 & 0.304 & 0.796 & 0.772 & 0.594 \\
        VoroMQA \citeyearpar{olechnovivc2017voromqa} & 0.412 & 0.419 & 0.291 & 0.688 & 0.651 & 0.505 \\
        RWplus \citeyearpar{zhang2010novel} & 0.192 & 0.167 & 0.137 & 0.033 & 0.056 & 0.011 \\
        SBROD \citeyearpar{karasikov2019smooth} & 0.431 & 0.413 & 0.291 & 0.551 & 0.569 & 0.393 \\
        Ornate \citeyearpar{pages2019protein} & 0.393 & 0.371 & 0.256 & 0.625 & 0.669 & 0.481 \\
        DimeNet \citeyearpar{klicpera2020directional} & 0.302 & 0.351 & 0.285 & 0.614 & 0.625 & 0.431 \\
        GraphQA \citeyearpar{baldassarre2021graphqa} & 0.357 & 0.379 & 0.251 & 0.821 & 0.820 & 0.618 \\
        PaiNN \citeyearpar{baldassarre2021graphqa} & 0.518 & 0.444 & 0.315 & 0.773 & 0.813 & 0.611 \\
        ET \citeyearpar{baldassarre2021graphqa} & 0.564 & 0.466 & 0.330 & 0.813 & 0.814 & 0.611 \\
        GVP \citeyearpar{jing2021equivariant} & 0.581 & 0.462 & 0.331 & 0.805 & 0.811 & 0.616 \\
        GBP \citeyearpar{10.1145/3534678.3539441} & \underline{0.612} & \underline{0.517} & \underline{0.372} & \underline{0.856} & \underline{0.853} & \underline{0.656} \\ \midrule
        \textsc{GCPNet} w/o Frames & 0.588 & 0.512 & 0.367 & 0.854 & 0.851 & 0.657 \\
        \textsc{GCPNet} w/o \textsc{ResGCP} & 0.576 & 0.509 & 0.365 & 0.852 & 0.847 & 0.648 \\
        \textsc{GCPNet} w/o Scalars & N/A & N/A & N/A & N/A & N/A & N/A \\
        \textsc{GCPNet} w/o Vectors & 0.571 & 0.497 & 0.356 & 0.802 & 0.804 & 0.608 \\
        \textsc{GCPNet} & \textbf{0.616} & \textbf{0.534} & \textbf{0.385} & \textbf{0.871} & \textbf{0.869} & \textbf{0.676} \\ \midrule
    \end{tabular}
\end{table}

\textbf{Evaluating ranking predictions for protein structure decoys.} Protein structure ranking (PSR) requires methods to predict the overall quality of a 3D protein structure when comparing it to a reference (i.e., native) protein structure \cite{townshend2020atom3d}. The quality of a protein structure is reported as a single scalar value representing a method's predicted global distance test (GDT\_TS) score \cite{zemla2003lga} between the provided decoy structure and the native structure. Such information is crucial in drug discovery efforts when one is tasked with designing a drug (e.g., ligand) that should bind to a particular protein target, notably when such targets have not yet had their 3D structures experimentally determined and have rather had them predicted computationally using methods such as AlphaFold 2 \cite{jumper2021highly}. The respective dataset for this SE(3)-invariant task is also derived from the ATOM3D dataset \cite{townshend2020atom3d} and is comprised of 40,950 decoy structures corresponding to 649 total targets, where cross-validation splits are created according to a target's release year in the Critical Assessment of Techniques for Protein Structure Prediction (CASP) competition \cite{kryshtafovych2021critical}. Results are reported in terms of the Pearson's correlation ($p$), Spearman's correlation ($Sp$), and Kendall's tau correlation ($K$) between a method's predictions on the test dataset and the corresponding ground-truth GDT\_TS values, where local results are averaged across predictions for individual targets and global results are averaged directly across all targets. Baseline comparison methods for this task include a composition of state-of-the-art CNNs, GNNs, and ENNs, as well as previous statistics-based methods.

Conveying a similar message to that in Table \ref{table:lba_results}, the results in Table \ref{table:psr_results} demonstrate that, in operating on atom-level protein graphs, \textsc{GCPNet} performs best against all other state-of-the-art models for the task of estimating a 3D protein structure's quality (i.e., PSR). In this setting, \textsc{GCPNet} outperforms all other methods across all local and global metrics by 2.5\% on average. Once again, \textsc{GCPNet}'s predictions are highly statistically significant, this time with Pearson, Spearman, and Kendall tau p-values all below $1e-50$, respectively.

\textbf{Identifying components for effective protein structure ranking.} Our ablation studies with \textsc{GCPNet}, in the context of PSR, once more reveal that the design of our local frames, \textsc{ResGCP} module, and scalar and vector feature channels are all beneficial for enhancing \text{GCPNet}'s ability to analyze a given 3D graph input. Here, in sensitizing the model to chemical chirality, our local frame embeddings improve \textsc{GCPNet}'s performance for PSR by 4\% on average. Similarly, our \textsc{ResGCP} module improves the model's performance by 5\%. Interestingly, without access to scalar-valued node and edge features, \textsc{GCPNet} is unable to produce valid predictions for the PSR test dataset due to what appears to be a phenomenon of vector-wise latent variable collapse \cite{pmlr-v89-dieng19a}. This finding suggests that, for the PSR task, the baseline \textsc{GCPNet} model relies strongly on the scalar-valued representations it produces. Lastly, including vector-valued features within \textsc{GCPNet} improves the model's performance for the PSR task by 9\%.

\begin{table}[t]
    \caption{Comparison of \textsc{GCPNet} with baseline methods for the NMS task. Results are reported in terms of the MSE for future position prediction over four test datasets of increasing modeling difficulty, graph sizes, and composed force field complexities. The final column reports each method's MSE averaged across all four test datasets. The best results for this task are in \textbf{bold}, and the second-best results are \underline{underlined}. N/A denotes an experiment that could not be performed due to a method's numerical instability.}
    \label{table:nms_results}
    \centering
    % \tiny  % Or footnotesize, scriptsize, tiny, etc.
    \begin{tabular}{lccccc}
        & & & & & \\ \midrule
        Method & ES(5) & ES(20) & G+ES(20) & L+ES(20) & Average \\ \midrule
        GNN \citeyearpar{pmlr-v162-du22e} & 0.0131 & 0.0720 & 0.0721 & 0.0908 & 0.0620 \\
        TFN \citeyearpar{pmlr-v162-du22e} & 0.0236 & 0.0794 & 0.0845 & 0.1243 & 0.0780 \\
        SE(3)-Transformer \citeyearpar{pmlr-v162-du22e} & 0.0329 & 0.1349 & 0.1000 & 0.1438 & 0.1029 \\
        Radial Field \citeyearpar{pmlr-v162-du22e} & 0.0207 & 0.0377 & 0.0399 & 0.0779 & 0.0441 \\
        PaiNN \citeyearpar{schutt2021equivariant} & 0.0158 & N/A & N/A & N/A & N/A \\
        ET \citeyearpar{tholke2022equivariant} & 0.1653 & 0.1788 & 0.2122 & 0.2989 & 0.2138 \\
        EGNN \citeyearpar{pmlr-v162-du22e} & 0.0079 & 0.0128 & 0.0118 & 0.0368 & 0.0173 \\
        ClofNet \citeyearpar{pmlr-v162-du22e} & \textbf{0.0065} & \underline{0.0073} & \textbf{0.0072} & 0.0251 & 0.0115 \\ \midrule
        \textsc{GCPNet} w/o Frames & \underline{0.0067} & 0.0074 & 0.0074 & \underline{0.0200} & \underline{0.0103} \\
        \textsc{GCPNet} w/o \textsc{ResGCP} & 0.0090 & 0.0135 & 0.0099 & 0.0278 & 0.0150 \\
        \textsc{GCPNet} w/o Scalars & 0.0119 & 0.0173 & 0.0170 & 0.0437 & 0.0225 \\
        \textsc{GCPNet} & 0.0070 & \textbf{0.0071} & \underline{0.0073} & \textbf{0.0173} & \textbf{0.0097} \\ \midrule
    \end{tabular}
\end{table}

\textbf{Evaluating trajectory predictions for Newtonian many-body systems.} Newtonian many-body systems modeling (NMS) asks methods to forecast the future positions of particles in many-body systems of various sizes \cite{pmlr-v162-du22e}, bridging the gap between the domains of machine learning and physics. In our experimental results for the NMS task, the four systems (i.e., datasets) on which we evaluate each method are comprised of increasingly more nodes and are influenced by force fields of increasingly complex directional origins for which to model, namely electrostatic force fields for 5-body (ES(5)) and 20-body (ES(20)) systems as well as for 20-body systems under the influence of an additional gravity field (G+ES(20)) and Lorentz-like force field (L+ES(20)), respectively. The four datasets for this SE(3)-equivariant task were generated using the descriptions and source code of \cite{pmlr-v162-du22e}, where each dataset is comprised of 7,000 total trajectories. Results are reported in terms of the mean squared error (MSE) between a method's node position predictions on the test dataset and the corresponding ground-truth node positions after 1,000 timesteps. Baseline comparison methods for this task include a collection of state-of-the-art GNNs, ENNs, and Transformers.

The results in Table \ref{table:nms_results} show that \textsc{GCPNet} achieves the lowest MSE averaged across all four NMS datasets, improving upon the state-of-the-art MSE for trajectory predictions in this task by 19\% on average. In particular, \textsc{GCPNet} achieves the best results for two of the four NMS datasets considered in this work, where these two datasets are respectively the first and third most difficult NMS datasets for methods to model. On the two remaining datasets, \textsc{GCPNet} matches the performance of prior state-of-the-art methods. Moreover, across all four datasets, \textsc{GCPNet}'s trajectory predictions yield an RMSE of 0.0963 and achieve Pearson, Spearman, and Kendall's tau correlations of 0.999, 0.999, and 0.981, respectively, where all such correlation values are highly statistically significant (i.e., p-values $< 1e-50$). Note that, to calculate these correlation values, we score \textsc{GCPNet}'s vector-valued predictions independently for each coordinate axis and then average the resulting metrics.

\textbf{Analyzing components for successful trajectory forecasting.} Once again, our ablation studies with \textsc{GCPNet} demonstrate the importance of \textsc{GCPNet}'s local frame embeddings, scalar information, and \textsc{ResGCP} module. Here, we note that we were not able to include an ablation study on \textsc{GCPNet}'s vector-valued features since they are directly used to predict node position displacements for trajectory forecasting. Table \ref{table:nms_results} shows that each model component synergistically enables \textsc{GCPNet} to achieve new state-of-the-art results for the NMS task. In enabling the model to detect global forces, our proposed local frame embeddings improve \textsc{GCPNet}'s ability to learn many-body system dynamics by 6\% on average across all dataset contexts. Specifically interesting to note is that these local frame embeddings improve the model's trajectory predictions within the most complex dataset context (i.e., L+ES(20)) by 14\%, suggesting that such frame embeddings improve \textsc{GCPNet}'s ability to learn many-body system dynamics even in the presence of complex global force fields. Furthermore, \textsc{GCPNet}'s \textsc{ResGCP} module and scalar-valued features improve the model's performance for modeling many-body systems by 35\% and 57\%, respectively.

Across all tasks studied in this work, \textsc{GCPNet} improves upon the overall performance of all previous methods. Our experiments demonstrate this for both node-level (e.g., NMS) and graph-level (e.g., LBA) prediction tasks, verifying \textsc{GCPNet}'s ability to encode useful information for both scales of granularity. Furthermore, we have demonstrated the importance of each model component within \textsc{GCPNet}, showing how these components are complementary to each other in the context of representation learning over 3D molecular data. We will now proceed to describe the design and operations of our proposed \textsc{GCPNet} model architecture.

\section{Methods}
\subsection{Preliminaries}

\subsubsection{Overview of the Problem Setting}
We represent a 3D molecular structure as a 3D $k$-nearest neighbors ($k$-NN) graph $\mathcal{G} = (\mathcal{V}, \mathcal{E})$ with $\mathcal{V}$ and $\mathcal{E}$ representing the graph's set of nodes and set of edges, respectively, and $N = |\mathcal{V}|$ and $E = |\mathcal{E}|$ representing the number of nodes and the number of edges in the graph, respectively. In addition, $\mathbf{X} \in \mathbb{R}^{N \times 3}$ represents the respective Cartesian coordinates for each node. We then design E(3)-invariant (i.e., 3D rotation, reflection, and translation-invariant) node features $\mathbf{H} \in \mathbb{R}^{N \times h}$ and edge features $\mathbf{E} \in \mathbb{R}^{E \times e}$ as well as O(3)-equivariant (3D rotation and reflection-equivariant) node features $\bm{\chi} \in \mathbb{R}^{N \times (m \times 3)}$ and edge features $\bm{\xi} \in \mathbb{R}^{E \times (x \times 3)}$, respectively.

Upon constructing such features, we apply several layers of graph message-passing using a neural network $\bm{\Phi}$ (which later on we refer to as \textsc{GCPNet}) that updates node and edge features using invariant and equivariant representations for the corresponding feature types. Importantly, $\bm{\Phi}$ guarantees, by design, \textit{SE(3) equivariance} with respect to its vector-valued input coordinates and features (i.e., $x_{i} \in \mathbf{X}$, $\chi_{i} \in \bm{\chi}$, and $\xi_{ij} \in \bm{\xi}$) and \textit{SE(3)-invariance} regarding its scalar features (i.e., $h_{i} \in \mathbf{H}$ and $e_{ij} \in \mathbf{E}$). In addition, $\bm{\Phi}$'s scalar graph representations achieve \textit{geometric self-consistency} for the 3D structure of the input molecular graph $\mathcal{G}$, sensitizing them to the effects of molecular chirality while making them uniquely identifiable under 3D rotations. Lastly, geometric completeness requires methods that accept 3D molecular graph inputs to be able to discern the local geometric environment of a given atom with no directional ambiguities. This enables geometry-complete methods such as $\bm{\Phi}$ to detect the presence and influence of global force fields acting on the graph inputs. We formalize these equivariance, geometric self-consistency, and geometric completeness constraints using the three following definitions, where $\Box^{\prime}$ represents a feature that has been updated by our neural network.
\newpage
\begin{definition}
    \label{definition:1}
    (SE(3) Equivariance)\textbf{.} \\ \\
    Given $(\mathbf{H}^{\prime}, \mathbf{E}^{\prime}, \mathbf{X}^{\prime}, \bm{\chi}^{\prime}, \bm{\xi}^{\prime}) = \bm{\Phi}(\mathbf{H}, \mathbf{E}, \mathbf{X}, \bm{\chi}, \bm{\xi})$, we have\\
    $(\mathbf{H}^{\prime}, \mathbf{E}^{\prime}, \mathbf{Q}\mathbf{X}^{\prime^{\mathrm{T}}} + \mathbf{g}, \mathbf{Q}\bm{\chi}^{\prime^{\mathrm{T}}}, \mathbf{Q}\bm{\xi}^{\prime^{\mathrm{T}}}) = \bm{\Phi}(\mathbf{H}, \mathbf{E}, \mathbf{Q}\mathbf{X}^{\mathrm{T}} + \mathbf{g}, \mathbf{Q}\bm{\chi}^{\mathrm{T}}, \mathbf{Q}\bm{\xi}^{\mathrm{T}})$, \\ $\quad \forall \mathbf{Q} \in \textit{SO}(3), \forall \mathbf{g} \in \mathbb{R}^{3 \times 1}$. 
\end{definition}
\begin{definition}
    \label{definition:2}
    (Geometric Self-Consistency)\textbf{.} \\ \\
    Given a pair of molecular graphs $\mathcal{G}_{1}$ and $\mathcal{G}_{2}$, \\
    with $\mathbf{X}^{1} = \{\mathbf{x}_{i}^{1}\}_{i = 1, ..., N}$ and $\mathbf{X}^{2} = \{\mathbf{x}_{i}^{2}\}_{i = 1, ..., N}$, respectively,\\ a geometric representation $\bm{\Phi}(\mathbf{H}, \mathbf{E}) = \bm{\Phi}(\mathcal{G})$ is considered \\ geometrically self-consistent if $\bm{\Phi}(\mathcal{G}^{1}) = \bm{\Phi}(\mathcal{G}^{2}) \Longleftrightarrow \exists \mathbf{Q} \in \textit{SO}(3), \exists \mathbf{g} \in \mathbb{R}^{3 \times 1}$, \\
    for $i = 1, ..., n, \mathbf{X}_{i}^{1^{\mathrm{T}}} = \mathbf{Q} \mathbf{X}_{i}^{2^{\mathrm{T}}} + \mathbf{g}$ \cite{wang2022comenet}.
\end{definition}
\begin{definition}
    \label{definition:3}
    (Geometric Completeness)\textbf{.} \\ \\
    Given a positional pair of nodes $(x_{i}^{t}, x_{j}^{t})$ in a 3D graph $\mathcal{G}$,\\
    with vectors $a_{ij}^{t} \in \mathbb{R}^{1 \times 3}$, $b_{ij}^{t} \in \mathbb{R}^{1 \times 3}$, and $c_{ij}^{t} \in \mathbb{R}^{1 \times 3}$ derived from $(x_{i}^{t}, x_{j}^{t})$, \\
    a local geometric representation $\bm{\mathcal{F}}_{ij}^{t} = (a_{ij}^{t}, b_{ij}^{t}, c_{ij}^{t}) \in \mathbb{R}^{3 \times 3}$ is considered\\
    geometrically complete if $\bm{\mathcal{F}}_{ij}^{t}$ is non-degenerate, thereby forming a \\
    \textit{local orthonormal basis} located at the tangent space of $x_{i}^{t}$ \cite{pmlr-v162-du22e}.
\end{definition}
\subsubsection{SE(3)-equivariant complete representations}
Representation learning on 3D molecular structures is a challenging task for a variety of reasons: (1) an expressive representation learning model should be able to predict arbitrary vector-valued quantities for each atom and atom pair in the molecular structure (e.g., using $\bm{\chi}^{\prime}$ and $\bm{\xi}^{\prime}$ to predict side-chain atom positions and atom-atom displacements for each residue in a 3D protein graph); (2) arbitrary rotations or translations to a 3D molecular structure should affect only the vector-valued representations a model assigns to a molecular graph's nodes or edges, whereas such 3D transformations of the molecular structure should not affect the model's scalar representations for nodes and edges \cite{pmlr-v162-du22e}; (3) the geometrically invariant properties of a molecule's 3D structure should be uniquely identifiable by a model; and (4) in a geometry-complete manner, scalar and vector-valued representations should mutually exchange information between nodes and edges during a model's forward pass for a 3D input graph, as these information types can be correlatively related (e.g., a scalar feature such as the $L_{2}$ norm of a vector $v$ can be associated with the vector of origin $v$) \cite{10.1145/3534678.3539441, morehead2022geometric}.

In line with this reasoning, we need to ensure that the coordinates our model predicts for the node positions in a molecular graph $\mathcal{G}$ transform according to SE(3) transformations of the input positions. This runs in contrast to previous methods that remain strictly E(3)-equivariant or E(3)-invariant to 3D transformations of the input $\mathcal{G}$ and consequently ignore the important effects of molecular chirality. At the same time, the model should jointly update the scalar and vector-valued features of $\mathcal{G}$ according to their respective molecular symmetry groups to increase the model's expressiveness in approximating geometric and physical quantities \cite{brandstetter2021geometric}. To increase its generalization capabilities, the model should also disambiguate any \textit{geometric directions} within its local node environments and should maintain SE(3)-invariance of its scalar representations when the input graph is transformed in 3D space. Following \cite{wang2023learning}, this helps prevent the model from losing important geometric or chiral information (i.e., becoming geometrically self-inconsistent) during graph message-passing. One way to do this is to introduce a new type of message-passing neural network.

\begin{figure}[t]
\centering
\includegraphics[width=\textwidth]{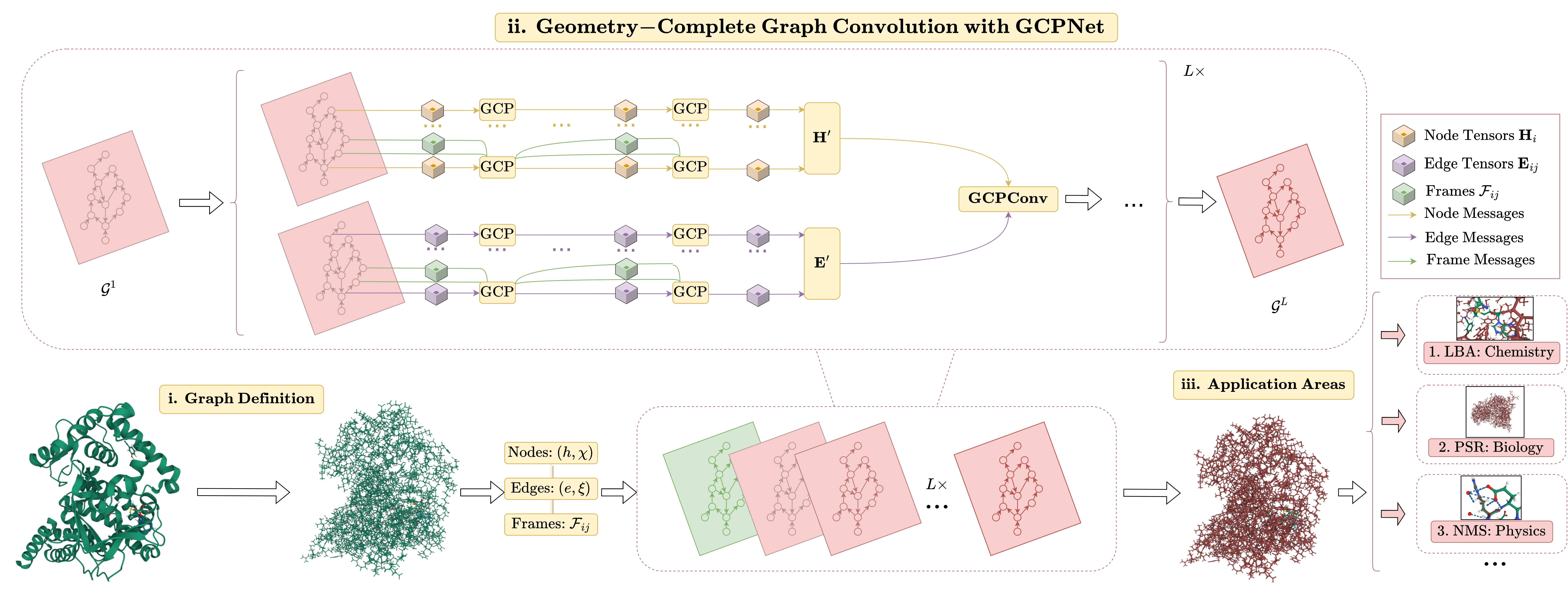}
\caption{A framework overview for our proposed \textit{Geometry-Complete Perceptron Network} (\textsc{GCPNet}). Our framework consists of (\textbf{i.}) a graph (topology) definition process, (\textbf{ii.}) a \textsc{GCPNet}-based graph neural network for 3D molecular representation learning, and (\textbf{iii.}) demonstrated application areas for \textsc{GCPNet}. Zoom in for the best viewing experience.}
\label{figure:gcpnet}
\end{figure}

\subsection{\textsc{GCPNet Model Architecture}}
\label{section:methodology}

\begin{figure}[t]
% \centering
\includegraphics[width=\columnwidth]{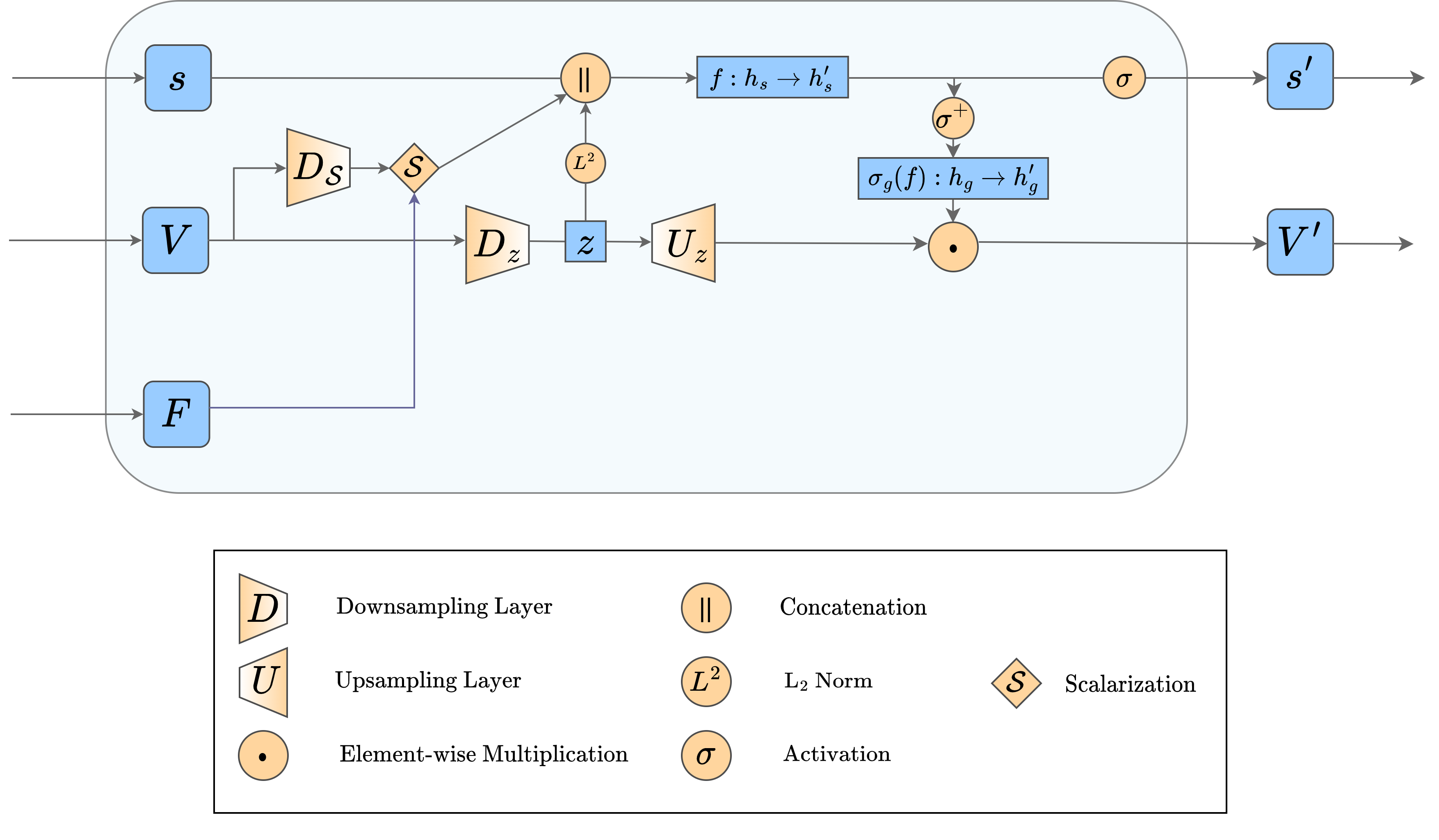} % Reduce the figure size so that it is slightly narrower than the column. Don't use precise values for figure width.This setup will avoid overfull boxes.
\caption{An overview of our proposed Geometry-Complete Perceptron ($\mathbf{GCP}$) module. The $\mathbf{GCP}$ module introduces node and edge-centric encodings of 3D frames as input features that are used to directly update both scalar and vector-valued features with geometric information-completeness guarantees as well as chirality sensitivity.}
\label{figure:gcp}
\end{figure}

Towards this end, we introduce our architecture for $\bm{\Phi}$ satisfying Defs. (\ref{definition:1}), (\ref{definition:2}), and (\ref{definition:3}) which we refer to as the Geometry-Complete SE(3)-Equivariant Perceptron Network (\textsc{GCPNet}). We illustrate the \textsc{GCPNet} algorithm in Figure \ref{figure:gcpnet} and outline it in Algorithm \ref{algorithm:1}. Subsequently, we expand on our definition for $\mathbf{GCP}$ and $\mathbf{GCPConv}$ in Sections \ref{section:gcp_module} and \ref{section:gcpconv_layer}, respectively, while further illustrating $\mathbf{GCP}$ in Figure \ref{figure:gcp}.

We can then prove the following three propositions (see Appendices \ref{section:prop_1_proof} through \ref{section:prop_3_proof} for a more detailed description of the \textsc{GCPNet} algorithm and its equivariant properties).
\begin{itemize}
    \item \textbf{Proposition 1.} \textit{\textsc{GCPNets} are SE(3)-equivariant \\ $\longrightarrow$ Def. (\ref{definition:1}).}
    \item \textbf{Proposition 2.} \textit{\textsc{GCPNets} are geometry self-consistent \\ $\longrightarrow$ Def. (\ref{definition:2}).}
    \item \textbf{Proposition 3.} \textit{\textsc{GCPNets} are geometry-complete \\ $\longrightarrow$ Def. (\ref{definition:3}).}
\end{itemize}

\subsubsection{Geometry-Complete Perceptron Module}
\label{section:gcp_module}

As illustrated in Figure \ref{figure:gcp}, \textsc{GCPNet} represents the features for nodes within an input graph as a tuple $(h, \chi)$ to distinguish scalar features $(h \in \mathbb{R}^{h})$ from vector-valued features $(\chi \in \mathbb{R}^{m \times 3})$. Similarly, \textsc{GCPNet} represents an input graph's edge features as a tuple $(e, \xi)$ to differentiate scalar features $(e \in \mathbb{R}^{e})$ from vector-valued features $(\xi \in \mathbb{R}^{x \times 3})$. For conciseness, we will subsequently refer to both node and edge feature tuples as $(s, V)$. We then define $\mathbf{GCP}_{\mathcal{F}_{ij}, \lambda}(\cdot)$ to represent the $\mathbf{GCP}$ encoding process, where $\lambda$ represents a downscaling hyperparameter (e.g., 3) and $\bm{\mathcal{F}}_{ij} \in \mathbb{R}^{3 \times 3}$ denotes the SO(3)-equivariant (i.e., 3D rotation-equivariant) frames constructed using the $\mathbf{Localize}$ operation (i.e., the $\mathbf{EquiFrame}$ operation of \cite{pmlr-v162-du22e}) in Algorithm \ref{algorithm:1}. Specifically, the frame encodings are defined as $\bm{\mathcal{F}}_{ij}^{t} = (a_{ij}^{t}, b_{ij}^{t}, c_{ij}^{t})$, with $a_{ij}^{t} = \frac{x_{i}^{t} - x_{j}^{t}}{ \lVert x_{i}^{t} - x_{j}^{t} \rVert }, b_{ij}^{t} = \frac{x_{i}^{t} \times x_{j}^{t}}{ \lVert x_{i}^{t} \times x_{j}^{t} \rVert },$ and $c_{ij}^{t} = a_{ij}^{t} \times b_{ij}^{t}$, respectively. In Appendix \ref{section:prop_3_proof}, we discuss how these frame encodings are direction information-complete for edges, allowing networks incorporating them to effectively detect and leverage for downstream tasks the force fields present within real-world many-body systems such as small molecules and proteins.

\textit{Expressing Vector Representations with $V$.} The $\mathbf{GCP}$ module then expresses vector representations $V$ as follows. The features $V$ with representation depth $r$ are downscaled by $\lambda$.
\begin{equation}
    z = \{v \bm{w}_{d_{z}} | \bm{w}_{d_{z}} \in \mathbb{R}^{r \times (r / \lambda)}\}
\end{equation}
Additionally, $V$ is separately downscaled in preparation to be subsequently embedded as direction-sensitive edge scalar features.
\begin{equation}
    V_{s} = \{v \bm{w}_{d_{s}} | \bm{w}_{d_{s}} \in \mathbb{R}^{r \times (3 \times 3)}\}
\end{equation}
\textit{Deriving Scalar Representations $s^{\prime}$.} To update scalar representations, the $\mathbf{GCP}$ module, in the following manner, derives two invariant sources of information from $V$ and combines them with $s$:
\begin{equation}
\label{eq:3}
    q_{ij} = (V_{s} \cdot \mathcal{F}_{ij}) \in \mathbb{R}^{9}
\end{equation}
\begin{align}
\label{eq:4}
    q = \begin{cases} \frac{1}{\lvert \mathcal{N}(i) \rvert} \sum_{j \in \mathcal{N}(i)} q_{ij} & \text{if $V_{s}$ represents nodes} \\
    q_{ij} & \text{if $V_{s}$ represents edges} \end{cases}
\end{align}
\begin{equation}
\label{eq:5}
    s_{(s, q, z)} = s \cup q \cup \lVert z \rVert_{2}
\end{equation}
where $\cdot$ denotes the inner product, $\mathcal{N}(\cdot)$ represents the neighbors of a node, and $\lVert \cdot \rVert_{2}$ denotes the $L_{2}$ norm. Then, denote $t$ as the representation depth of $s$, and let $s_{(s, q, z)} \in \mathbb{R}^{t + 9 + (r / \lambda)}$ with representation depth $(t + 9 + (r / \lambda))$ be projected to $s^{\prime}$ with representation depth $t^{\prime}$:
\begin{equation}
    s_{v} = \{s_{(s, q, z)} \bm{w}_{s} + \bm{b}_{s} | \bm{w}_{s} \in \mathbb{R}^{(t + 9 + (r / \lambda)) \times t^{\prime}} \}
\end{equation}
\begin{equation}
    s^{\prime} = \sigma_{s}(s_{v})
\end{equation}

Note that embedding geometric frames $\mathcal{F}_{ij}$ as $q_{ij}$ in Equation \ref{eq:3} ultimately enables \textsc{GCPNet} to iteratively learn chirality-sensitive and global force-aware representations of each 3D network input. Moreover, Equation \ref{eq:4} allows \textsc{GCPNet} to encode local geometric substructures for each node, where the theoretical importance of such network behavior is discussed in detail by \cite{du2023new}.

\textit{Deriving Vector Representations $V^{\prime}$.} The $\mathbf{GCP}$ module then concludes by updating vector representations as follows:
\begin{equation}
    V_{u} = \{z \bm{w}_{u_{z}} | \bm{w}_{u_{z}} \in \mathbb{R}^{(r / \lambda) \times r^{\prime}}\}
\end{equation}
\begin{equation}
    V^{\prime} = \{V_{u} \odot \sigma_{g}(\sigma^{+}(s_{v}) \bm{w}_{g} + \bm{b}_{g}) | \bm{w}_{g} \in \mathbb{R}^{t^{\prime} \times r^{\prime}} \}
\end{equation}
where $\odot$ represents element-wise multiplication and the gating function $\sigma_{g}$ is applied row-wise to preserve SO(3) equivariance within $V^{\prime}$.

Conceptually, the $\mathbf{GCP}$ module is autoregressively applied to tuples $(s, V)$ a total of $\omega$ times to derive rich scalar and vector-valued features. The module does so by blending both feature types iteratively with the 3D direction and information completeness guarantees provided by geometric frame encodings $\bm{\mathcal{F}}_{ij}$. We note that this model design runs in contrast with prior graph neural networks for physical systems such as GVP-GNNs \cite{pmlr-v162-du22e} and ClofNet \cite{jing2020learning}, which are either insensitive to chemical chirality and global atomic forces or do not directly learn geometric features for downstream prediction tasks, making the proposed $\mathbf{GCP}$ module well suited for learning directly from 3D molecular graphs.

\subsection{Learning from 3D Graphs with \textsc{GCPNet}}
In this section, we propose a flexible manner in which to perform 3D graph convolution with our proposed $\mathbf{GCP}$ module, as illustrated in Figure \ref{figure:gcpnet} and employed in Algorithm \ref{algorithm:1}.

\subsubsection{Geometry-Complete Graph Convolution.}
\label{section:gcpconv_layer}
Let $\mathcal{N}$(i) denote the neighbors of node $n_{i}$, selected using a distance-based metric such as k-nearest neighbors or a radial distance cutoff. Subsequently, we define a single layer $l$ of geometry-complete graph convolution as
\begin{equation}
\label{eq:10}
n_{i}^{l} = \phi^{l}(n_{i}^{l - 1}, \mathcal{A}_{\forall j \in \mathcal{N}(i)} \Omega_{\omega}^{l}(n_{i}^{l - 1}, n_{j}^{l - 1}, e_{ij}, \mathcal{F}_{ij})),
\end{equation}
where $n_{i}^{l} = (h_{i}^{l}, \chi_{i}^{l})$; $e_{ij}=(e_{ij}^{0}, \xi_{ij}^{0})$; $\bm{\Phi}$ is a trainable function denoted as $\mathbf{GCPConv}$; $l$ signifies the representation depth of the network; $\mathcal{A}$ is a permutation-invariant aggregation function; and $\Omega_{\omega}$ represents a message-passing function corresponding to the $\omega$-th $\mathbf{GCP}$ message-passing layer. We proceed to expand on the operations of each graph convolution layer as follows.

To start, messages between source nodes $i$ and neighboring nodes $j$ are first constructed as
\begin{equation}
    m_{ij}^{0} = \mathbf{GCP}(n_{i}^{0} \cup n_{j}^{0} \cup e_{ij}, \mathcal{F}_{ij})
\end{equation}
where $\cup$ denotes a concatenation operation. Then, up to the $\omega$-th iteration, each message is updated by the $m$-th message update layer using residual connections as
\begin{align}
    \Omega_{\omega}^{l} &= \mathbf{ResGCP}_{\omega}^{l}(m_{ij}^{l - 1}, \mathcal{F}_{ij}), \\
    \mathbf{ResGCP}_{\eta}^{l}(z_{i}^{l - 1}, \mathcal{F}_{ij}) &= z_{i}^{l - 1} + \mathbf{GCP}_{\eta}^{l}(z_{i}^{l - 1}, \mathcal{F}_{ij}),
\end{align}
where we empirically find such residual connections between message representations to reduce oversmoothing within \textsc{GCPNet} by mitigating the problem of vanishing gradients.

Updated node features $\hat{n}^{l}$ are then derived residually using an aggregation of generated messages as
\begin{equation}
    \hat{n}^{l} = n^{l - 1} + f(\{\Omega_{\omega, v_{i}}^{l} | v_{i} \in \mathcal{V}\}),
\end{equation}
where $f$ represents an aggregation function such as a summation or mean that is invariant to permutations of node ordering. The residual connection between $\hat{n}^{l}$ and $n^{l}$ is established here to encourage the network to update the representation space of node features in a layer-asynchronous manner.

To encourage $\textsc{GCPNet}$ to make its node feature representations independent of the size of each input graph, we then employ a node-centric feed-forward network to update node representations. Specifically, we apply to $\hat{n}^{l}$ a linear $\mathbf{GCP}$ function with shared weights $\phi_{f}$ followed by $r$ $\mathbf{ResGCP}$ modules, operations concisely portrayed as
\begin{align}
\label{eq:16}
    \Tilde{n}_{r - 1}^{l} &= \phi_{f}^{l}(\hat{n}^{l}) \\
    n^{l} &= \mathbf{ResGCP}_{r}^{l}(\Tilde{n}_{r - 1}^{l}).
\end{align}
Lastly, if one desires to update the positions of each node in $\mathcal{G}$ (e.g., as we do for tasks involving position-related predictions such as NMS), we propose a flexible, SE(3)-equivariant method to do so using a dedicated $\mathbf{GCP}$ module as follows:
\begin{align}
\label{eq:18}
    (h_{p_{i}}^{l}, \chi_{p_{i}}^{l}) &= \mathbf{GCP}_{p}^{l}(n_{i}^{l}, \mathcal{F}_{ij}) \\
    x_{i}^{l} &= x_{i}^{l - 1} + \chi_{p_{i}}^{l}, \mbox{where } \chi_{p_{i}}^{l} \in \mathbb{R}^{1 \times 3}.
\end{align}

\subsubsection{The \textsc{GCPNet} Algorithm}

\begin{algorithm}[tb]
\caption{\textsc{GCPNet}}
\label{algorithm:1}
    \begin{algorithmic}[1]
        \STATE {\bfseries Input:} $(h_{i} \in \mathbf{H}$, $\chi_{i} \in \bm{\chi})$, $(e_{ij} \in \mathbf{E}$, $\xi_{ij} \in \bm{\xi})$, \\ $x_{i} \in \mathbf{X}$, graph $\mathcal{G}$
        \STATE Initialize $\mathbf{X}^{0} = \mathbf{X}^{C} \leftarrow \mathbf{Centralize}(\mathbf{X})$
        \STATE $\bm{\mathcal{F}}_{ij} = \mathbf{Localize}(x_{i} \in \mathbf{X}^{0}, x_{j} \in \mathbf{X}^{0})$
        \STATE Project $(h_{i}^{0}$, $\chi_{i}^{0})$, $(e_{ij}^{0}$, $\xi_{ij}^{0}) \leftarrow \mathbf{GCP}_{e}((h_{i}$, $\chi_{i})$, $(e_{ij}$, $\xi_{ij})$, $\mathcal{F}_{ij})$
        \FOR{$l=1$ {\bfseries to} $L$}
        \STATE $(h_{i}^{l}$, $\chi_{i}^{l})$, $x_{i}^{l} = \mathbf{GCPConv}^{l}((h_{i}^{l - 1}$, $\chi_{i}^{l - 1})$, $(e_{ij}^{0}$, $\xi_{ij}^{0})$, $x_{i}^{l - 1}$, $\mathcal{F}_{ij})$
        \ENDFOR
        \IF{Updating Node Positions}
        \STATE $\bm{\mathcal{F}}_{ij}^{L} = \mathbf{Localize}(x_{i} \in \mathbf{X}^{l}$, $x_{j} \in \mathbf{X}^{l})$
        \STATE Finalize $(\mathbf{X}^{L}) \leftarrow \mathbf{Decentralize}(\mathbf{X}^{l})$
        \ELSE
        \STATE $x_{i}^{L} = x_{i}^{0}$
        \ENDIF
        \STATE Project $(h_{i}^{L}$, $\chi_{i}^{L})$, $(e_{ij}^{L}$, $\xi_{ij}^{L}) \leftarrow \mathbf{GCP}_{p}((h_{i}^{l}$, $\chi_{i}^{l})$, $(e_{ij}^{0}$, $\xi_{ij}^{0})$, $\mathcal{F}_{ij}^{L})$
        \STATE {\bfseries Output:} $(h_{i}^{L}$, $\chi_{i}^{L})$, $(e_{ij}^{L}$, $\xi_{ij}^{L})$, $x_{i}^{L}$
    \end{algorithmic}
\end{algorithm}

In this section, we describe our overall learning algorithm driven by \textsc{GCPNet} (Algorithm \ref{algorithm:1}). We also discuss the rationale behind our design decisions for \textsc{GCPNet} and provide examples of use cases in which one might apply \textsc{GCPNet} for specific learning tasks.

On Line 2 of Algorithm \ref{algorithm:1}, the $\mathbf{Centralize}$ operation removes the center of mass from each node position in the input graph to ensure that such positions are subsequently 3D translation-invariant.

Thereafter, following \cite{pmlr-v162-du22e}, the $\mathbf{Localize}$ operation on Line 3 crafts translation-invariant and SO(3)-equivariant frame encodings $\bm{\mathcal{F}}_{ij}^{t} = (a_{ij}^{t}, b_{ij}^{t}, c_{ij}^{t})$. As described in more detail in Appendix \ref{section:prop_3_proof}, these frame encodings are chirality-sensitive and direction information-complete for edges, imbuing networks that incorporate them with the ability to more easily detect force field interactions present in many real-world atomic systems, as we demonstrate through corresponding experiments in Section \ref{section:results}.

Before applying any geometry-complete graph convolution layers, on Line 4 we use $\mathbf{GCP}_{e}$ to embed our input node and edge features into scalar and vector-valued values, respectively, while incorporating geometric frame information. Subsequently, in Lines 5-6, each layer of geometry-complete graph convolution is performed autoregressively via $\mathbf{GCPConv}^{l}$ starting from these initial node and edge feature embeddings, all while maintaining information flow originating from the geometric frames $\bm{\mathcal{F}}_{ij}$.

On Lines 8 through 12, we finalize our procedure with which to update in an SE(3)-equivariant manner the position of each node in an input 3D graph. In particular, we update node positions by residually adding learned vector-valued node features ($\chi_{v_{i}}^{l}$) to the node positions produced by the previous $\mathbf{GCPConv}$ layer ($l - 1$). As shown in Appendix \ref{section:prop_1_proof}, such updates are initially SO(3)-equivariant, and on Line 10 we ensure these updates also become 3D translation-equivariant by adding back to each node position the input graph's original center of mass via the $\mathbf{Decentralize}$ operation. In total, this procedure produces SE(3)-equivariant updates to node positions. Additionally, for models that update node positions, we note that Line 9 updates frame encodings $\bm{\mathcal{F}}_{ij}$ using the model's final predictions for node positions to provide more information-rich feature projections on Line 14 via $\mathbf{GCP}_{p}$ to conclude the forward pass of \textsc{GCPNet}.

\subsubsection{Network Utilities.}
In summary, \textsc{GCPNet} receives an input 3D graph $\mathcal{G}$ with node positions $\bm{x}$, scalar node and edge features, $h$ and $e$, as well as vector-valued node and edge features, $\chi$ and $\xi$. The model is then capable of e.g., (1) predicting scalar node, edge, or graph-level properties while maintaining SE(3) invariance; (2) estimating vector-valued node, edge, or graph-level properties while ensuring SE(3) equivariance; or (3) updating node positions in an SE(3)-equivariant manner.

\section{Discussion}\label{section:discussion}

In light of our impressive results with \textsc{GCPNet}, future work on the model could involve researching more computationally-efficient variations of \textsc{GCPNet} that require fewer \textsc{GCP} message-passing layers within each \textsc{GCP} convolution layer or that embed geometric frames sparsely rather than in each \textsc{GCP} layer. Interestingly, \cite{du2023new} recently showed that \textsc{GCPNet}, in particular, is theoretically robust in terms of its geometric expressiveness of local geometric substructures and global geometric interaction terms. Nonetheless, \cite{du2023new} also highlight future directions for improving such methods' expressiveness for learning equivariant-valued features including incorporating higher-order equivariant tensors into one's message-passing procedure. Enhancing its geometric expressiveness to thereby increase its runtime efficiency would allow \textsc{GCPNet} to be used increasingly in new scientific and deep learning applications requiring high computational throughput (e.g., virtual screening of new drugs).

\section{Conclusion}\label{section:conclusion}

In this work, we introduced \textsc{GCPNet}, a state-of-the-art GNN for 3D molecular graph representation learning. We have demonstrated its utility through several benchmark studies. In future work, we aim to develop extensions of \textsc{GCPNet} that increase its geometric expressiveness as well as explore applications of \textsc{GCPNet} for generative modeling of molecular structures.

\bmhead{Code Availability}
The source code for \textsc{GCPNet} is available at \href{https://github.com/BioinfoMachineLearning/GCPNet}{https://github.com/BioinfoMachineLearning/GCPNet}.

\bmhead{Data Availability}
The data required to train new \textsc{GCPNet} models or reproduce our results for the NMS task are available under a Creative Commons Attribution 4.0 International Public License at \href{https://zenodo.org/record/7293186}{https://zenodo.org/record/7293186}. The data required to train new \textsc{GCPNet} models or reproduce our results for the RS task are available under an MIT License at \href{https://figshare.com/s/e23be65a884ce7fc8543}{https://figshare.com/s/e23be65a884ce7fc8543}. All other data required to train new \textsc{GCPNet} models or reproduce our results for the remaining tasks are available to download using our model training scripts available at \href{https://github.com/BioinfoMachineLearning/GCPNet}{https://github.com/BioinfoMachineLearning/GCPNet}.

\bmhead{Acknowledgments}
This work is partially supported by two NSF grants (DBI1759934 and IIS1763246), two NIH grants (R01GM093123 and R01GM146340), three DOE grants (DE-AR0001213, DE-SC0020400, and DE-SC0021303), and the computing allocation on the Summit compute cluster provided by the Oak Ridge Leadership Computing Facility.

\bmhead{Author Contributions Statement}
AM and JC conceived the project. AM designed the experiments. AM performed the experiments and collected the data. AM analyzed the data. AM and JC wrote the manuscript. AM and JC edited the manuscript.

\bmhead{Competing Interests Statement}
The authors declare no competing interests.

\backmatter


\bibliography{GCPNet}% bib file

\newpage

\begin{appendices}

\section{Proofs.}

\subsection{Proof of Proposition 1.}
\label{section:prop_1_proof}

\textit{Proof.} Suppose the vector-valued features given to the corresponding $\mathbf{GCPConv}$ layers in \textsc{GCPNet} are node features $\chi_{i}$ and edge features $\xi_{ij}$ that are O(3)-equivariant (i.e., 3D rotation and reflection-equivariant) by way of their construction. Additionally, suppose the scalar-valued features given to the respective $\mathbf{GCPConv}$ layers in \textsc{GCPNet} are E(3)-invariant (i.e., 3D rotation, reflection, and translation-invariant) node features $h_{i}$ and edge features $e_{ij}$.

\textbf{Translation equivariance.} In line with \cite{pmlr-v162-du22e}, the $\mathbf{Centralize}$ operation on Line 2 of Algorithm \ref{algorithm:1} first ensures that $\mathbf{X}^{0}$ becomes 3D translation invariant by the following procedure. Let $\mathbf{X}(t) = (\mathbf{x}_{1}(t), ..., \mathbf{x}_{n}(t))$ represent a many-body system at time $t$, where the centroid of the system is defined as
\begin{equation}
    C(t)=\frac{\mathbf{x}_{1}(t) + ... + \mathbf{x}_{n}(t)}{n}.
\end{equation}
Note that in uniformly translating the position of the system by a vector $\mathbf{v}$, we have $\mathbf{X}(t) + \mathbf{v} \longrightarrow C(t) + \mathbf{v}$, meaning that the centroid of the system translates in the same manner as the system itself. However, note that if at time $t = 0$ we recenter the origin of $\mathbf{X}$ to its centroid, we have
\begin{align*}
    \mathbf{X}(t) - C(0) \xrightarrow{\mbox{translation by } \mathbf{v} \mbox{ at } t = 0} \mathbf{X}(t) - C(0)
\end{align*}
which implies the system $\mathbf{X}$ is translation-\underline{invariant} under the centralized reference $\mathbf{X}(t) - C(0)$ when the translation vector $\mathbf{v}$ is applied to $\mathbf{X}$ at time $t = 0$. Concretely, in the case of translation-\underline{invariant} tasks such as predicting molecular properties or classifying point clouds, here we have successfully achieved 3D translation invariance. Moreover, for translation-\underline{equivariant} tasks such as forecasting the positions of a many-body system, we can achieve translation \underline{equivariance} by simply adding $C(0)$ back to the predicted positions. Therefore, using the above methodology, \textsc{GCPNets} are translation equivariant.

\textbf{Permutation equivariance.} Succinctly, we note that since \textsc{GCPNet} operates on graph-structured input data, permutation equivariance is guaranteed by design. For further discussion of why our proposed method as well as why other graph-based algorithms proposed previously are inherently permutation-equivariant, we refer readers to \cite{zaheer2017deep}. Therefore, \textsc{GCPNets} are permutation-equivariant.

\textbf{SO(3)-equivariant frames.} On Line 3 of Algorithm \ref{algorithm:1}, the $\mathbf{Localize}$ operation constructs SO(3)-equivariant (i.e., 3D rotation-equivariant) frames $\bm{\mathcal{F}}_{ij}$ in the following manner.

Define our frame encodings as
\begin{equation}
\label{eq:20}
    \bm{\mathcal{F}}_{ij}^{t} = (a_{ij}^{t}, b_{ij}^{t}, c_{ij}^{t}),
\end{equation}
where we have
\begin{equation}
\label{eq:21}
    a_{ij}^{t} = \frac{x_{i}^{t} - x_{j}^{t}}{ \lVert x_{i}^{t} - x_{j}^{t} \rVert }, b_{ij}^{t} = \frac{x_{i}^{t} \times x_{j}^{t}}{ \lVert x_{i}^{t} \times x_{j}^{t} \rVert }, c_{ij}^{t} = a_{ij}^{t} \times b_{ij}^{t}.
\end{equation}
The proof that $\bm{\mathcal{F}}_{ij}^{t}$ is equivariant under SO(3) transformations of its input space is included in \cite{pmlr-v162-du22e}. However, for completeness, we include a version of it here.

Let $g \in SO(3)$ be an action under which the positions in $X$ transform equivariantly, and $\bm{\mathcal{F}}_{ij}^{t}$ be defined as we have it in Equation \ref{eq:20} above. That is, we have
\begin{align*}
    (\mathbf{x}_{1}(t), ..., \mathbf{x}_{n}(t)) \xrightarrow{g} (g\mathbf{x}_{1}(t), ..., g\mathbf{x}_{n}(t)),
\end{align*}
where from the definition of $a_{ij}^{t}$ in Equation \ref{eq:21} we have
\begin{align*}
    a_{ij}^{t} \xrightarrow{g} g a_{ij}^{t}.
\end{align*}
Considering $b_{ij}^{t}$, from Equation \ref{eq:21} we have
\begin{align}
\label{eq:22}
    (g x_{i}(t)) \times (g x_{j}(t)) &= \det(g)(g^{T})^{-1}(x_{i}(t) \times x_{j}(t)) \notag\\ &= g(x_{i}(t) \times x_{j}(t)),
\end{align}
where using $g^{-1} = g^{T}$ for the orthogonal matrix $g$ gives us Equation \ref{eq:22}. Consequently, $b_{ij}^{t} \xrightarrow{g} g b_{ij}^{t}$. Lastly, by applying Equation \ref{eq:22} once again, we have that $c_{ij}^{t} \xrightarrow{g} g c_{ij}^{t}$.

Moreover, note that under reflections of $x$, we have $R : x \rightarrow -x$ which gives us $a_{ij}^{t} \rightarrow -a_{ij}^{t}$. Thereafter, by the right-hand rule, the cross product of two equivariant vectors gives us a pseudo-vector $b_{ij}^{t} = x_{i}^{t} \times x_{j}^{t} \rightarrow b_{ij}^{t}$, where subsequently it is implied that $c_{ij}^{t} \rightarrow -c_{ij}^{t}$. Consequently, we have $\det(-a_{ij}^{t}, b_{ij}^{t}, -c_{ij}^{t}) = 1$, informing us that the frame encodings $\bm{\mathcal{F}}_{ij}^{t}$ are \underline{not} reflection-equivariant (a symmetry that is important to not enforce when learning representations of chiral molecules such as proteins). Therefore, the frame encodings within \textsc{GCPNet} are SO(3)-equivariant.

Note, after the construction of these frames, that they are used on Line 4 of Algorithm \ref{algorithm:1} to embed all node and edge features (i.e., $h_{i}$, $e_{ij}$, $\chi_{i}$, and $\xi_{ij}$) using a single $\mathbf{GCP}$ module as well as in all subsequent $\mathbf{GCP}$ modules. We will now prove that the feature updates each $\mathbf{GCP}$ module makes with the frame encodings $\bm{\mathcal{F}}_{ij}^{t}$ defined in Equation \ref{eq:20} are SO(3)-equivariant.

\textbf{SO(3)-equivariant GCP module.} The operations of a $\mathbf{GCP}$ module are illustrated in Figure \ref{figure:gcp} and derived in Section \ref{section:gcp_module}. Their SO(3) invariance for scalar feature updates and SO(3) equivariance for vector-valued feature updates is proven as follows.

Following the proof of O(3) equivariance for the \textsc{GVP} module in \cite{jing2020learning}, the proof of SO(3) equivariance within the $\mathbf{GCP}$ module is similar, with the following modifications. Within the $\mathbf{GCP}$ module, the vector-valued features (processed separately for nodes and edges) are fed not only through a bottleneck block comprised of downward and upward projection matrices $\mathbf{D_{z}}$ and $\mathbf{U_{z}}$ but are also fed into a dedicated downward projection matrix $\mathbf{D_{\mathcal{S}}}$. The output of matrix multiplication between O(3)-equivariant vector features and $\mathbf{D_{\mathcal{S}}}$ yields O(3)-equivariant vector features $v_{i_{\mathcal{S}}}$ that are used as unique inputs for an SO(3)-invariant \underline{scalarization} operation. In particular, the following demonstrates the invariance of our design for matrix multiplication with our $\mathbf{GCP}$ module's projection matrices (e.g., $\mathbf{D_{\mathcal{S}}}$). Suppose $\mathbf{W}_{h} \in \mathbb{R}^{h \times v}$, $\mathbf{V} \in \mathbb{R}^{v \times 3}$, and $\mathbf{Q} \in SO(3) \in \mathbb{R}^{3 \times 3}$. In line with \cite{jing2020learning}, observe for $\mathbf{D} = (\mathbf{Q} \mathbf{V}^{\mathrm{T}}) \in \mathbb{R}^{3 \times v}$ that
\begin{align*}
    \lVert \mathbf{W}_{h} \mathbf{D}^{\mathrm{T}} \rVert_{2} = \lVert \mathbf{W}_{h} (\mathbf{V}^{\mathrm{T}})^{\mathrm{T}} \rVert_{2} = \lVert \mathbf{W}_{h} \mathbf{V} \rVert_{2}.
\end{align*}
Specifically, our SO(3)-invariant scalarization operation is defined as
\begin{equation}
\label{eq:23}
    q_{ij} = (v_{i_{\mathcal{S}}} \cdot a_{ij}^{t}, v_{i_{\mathcal{S}}} \cdot b_{ij}^{t}, v_{i_{\mathcal{S}}} \cdot c_{ij}^{t}),
\end{equation}
where $\bm{\mathcal{F}}_{ij}^{t} = (a_{ij}^{t}, b_{ij}^{t}, c_{ij}^{t})$ denotes the SO(3)-equivariant frame encodings defined in Equations \ref{eq:20} and \ref{eq:21}.

To prove that Equation \ref{eq:23} yields SO(3)-invariant scalar features, let $g \in SO(3)$ be an arbitrary orthogonal transformation. Then we have $v_{i_{\mathcal{S}}} \rightarrow g v_{i_{\mathcal{S}}}$, and similarly $\bm{\mathcal{F}}_{ij}^{t} = (a_{ij}^{t}, b_{ij}^{t}, c_{ij}^{t}) \rightarrow (g a_{ij}^{t}, g b_{ij}^{t}, g c_{ij}^{t})$. Now, similar to \cite{pmlr-v162-du22e}, we can derive that Equation \ref{eq:23} becomes
\begin{align}
    (v_{i_{\mathcal{S}}} \cdot a_{ij}^{t}, v_{i_{\mathcal{S}}} \cdot b_{ij}^{t}, v_{i_{\mathcal{S}}} \cdot c_{ij}^{t}) \notag\\ \rightarrow ((v_{i_{\mathcal{S}}})^{T} g^{T} g a_{ij}^{t}, (v&_{i_{\mathcal{S}}})^{T} g^{T} g b_{ij}^{t}, (v_{i_{\mathcal{S}}})^{T} g^{T} g c_{ij}^{t}) \notag\\ = (v_{i_{\mathcal{S}}} \cdot a_{ij}^{t}, v_{i_{\mathcal{S}}} \cdot b_{ij}^{t}&, v_{i_{\mathcal{S}}} \cdot c_{ij}^{t}),&
\end{align}
where we used the fact that $g^{T} g = I$ due to the orthogonality of $g$ (with $I$ being the identity matrix). Therefore, the scalarization operation proposed in Equation \ref{eq:23}, and previously in Equation \ref{eq:3} (in an alternative form), yields SO(3)-invariant scalars, which is in line with the results of \cite{qiao2022dynamic}.

The output of Equation \ref{eq:23}, $q_{ij}$, is then aggregated in Equation \ref{eq:4} and concatenated in Equation \ref{eq:5} with the $\mathbf{GCP}$ module's remaining O(3)-invariant scalar features (i.e., $L_{2}$ vector norm features). Note that introducing SO(3)-invariant scalar information into the $\mathbf{GCP}$ module in this way breaks the 3D reflection symmetry that previous geometric graph convolution modules enforced \cite{jing2020learning}, now giving rise within the $\mathbf{GCP}$ module to SO(3)-invariant and SO(3)-equivariant updates to scalar and vector-valued features, respectively. Therefore, scalar and vector-valued feature updates for nodes and edges within the $\mathbf{GCP}$ module are SO(3)-invariant and SO(3)-equivariant, respectively.

As in Section \ref{section:gcpconv_layer}, we now turn to discuss the operations within a single $\mathbf{GCPConv}$ layer, in particular proving that they maintain the respective SO(3) invariance and SO(3) equivariance for scalar and vector-valued features that the $\mathbf{GCP}$ module provides.

\textbf{SO(3)-equivariant GCPConv layer.} Via the corresponding proof in \cite{jing2020learning}, by way of induction all such operations in Equations \ref{eq:10}-\ref{eq:16} are respectively SE(3)-invariant and SO(3)-equivariant for features $m_{ij}^{l}=(m_{e_{ij}}^{l}, m_{\xi_{ij}}^{l})$. Thereby, so are features $n_{i}^{l}=(h_{i}^{l}, \chi_{i}^{l})$, given that the proof of equivariance for the equivariant $\mathbf{LayerNorm}$ and $\mathbf{Dropout}$ operations employed within each $\mathbf{GCPConv}$ has previously been concretized by \cite{jing2020learning}. Equation \ref{eq:18} concludes the operations of a single $\mathbf{GCPConv}$ layer by, as desired, updating the positions of each node $i$ in the 3D input graph. To do so, $\mathbf{GCPConv}$ residually updates current node positions $x_{i}^{l - 1}$ using SO(3)-equivariant vector-valued features $\chi_{p_{i}}^{l}$. Therefore, $\mathbf{GCPConv}$ layers are SO(3)-invariant for scalar feature updates and SO(3)-equivariant for vector-valued node position and feature updates.

\textbf{SE(3)-equivariant GCPNet.} Lastly, as desired, Line 10 of Algorithm \ref{algorithm:1} adds $C(0)$ back to the predicted node positions $\mathbf{X}^{l}$ as provided by each $\mathbf{GCPConv}$ layer, ultimately imbuing position updates within $\mathbf{X}^{l}$ with SE(3) equivariance. Line 14 then concludes \textsc{GCPNet} by using the latest frame encodings $\bm{\mathcal{F}}_{ij}^{t}$ to perform, as desired, a final SO(3)-invariant and SO(3)-equivariant projection for scalar and vector-valued features, respectively. Therefore, as desired, \textsc{GCPNets} are SE(3)-invariant for scalar feature updates, SE(3)-equivariant for vector-valued node position and feature updates, and, as a consequence, satisfy the constraint proposed in Def. \ref{definition:1}.

\qed

\subsection{Proof of Proposition 2.}
\label{section:prop_2_proof}

\textit{Proof.} The proof of SE(3) invariance for scalar node and edge features, $h_{i}$ and $e_{ij}$, follows as a corollary of Appendix \ref{section:prop_1_proof} (\textbf{SE(3)-equivariant GCPNet}). Therefore, \textsc{GCPNets} are SE(3)-invariant concerning their predicted scalar node and edge features and, as a consequence, are geometrically self-consistent according to the constraint in Def. \ref{definition:2}.

\qed

\subsection{Proof of Proposition 3.}
\label{section:prop_3_proof}

\textit{Proof.} Suppose that \textsc{GCPNet} designates its local geometric representation for layer $t$ to be $\bm{\mathcal{F}}_{ij}^{t} = (a_{ij}^{t}, b_{ij}^{t}, c_{ij}^{t})$, where $a_{ij}^{t} = \frac{x_{i}^{t} - x_{j}^{t}}{ \lVert x_{i}^{t} - x_{j}^{t} \rVert }, b_{ij}^{t} = \frac{x_{i}^{t} \times x_{j}^{t}}{ \lVert x_{i}^{t} \times x_{j}^{t} \rVert },$ and $c_{ij}^{t} = a_{ij}^{t} \times b_{ij}^{t}$, respectively. As in \cite{pmlr-v162-du22e}, this formulation of $\bm{\mathcal{F}}_{ij}^{t}$ is proven in Appendix \ref{section:prop_1_proof} (\textit{SO(3)-equivariant frames}) to be an SO(3)-equivariant local orthonormal basis at the tangent space of $x_{i}^{t}$ and is thereby geometrically complete. Note this implies that $\textsc{GCPNet}$ permits no loss of geometric information as discussed in Appendix A.5 of \cite{pmlr-v162-du22e}. Therefore, \textsc{GCPNets} are geometry-complete and satisfy the constraint proposed in Def. \ref{definition:3}.

\qed

\newpage

\section{Additional Experiments and Results.}
\label{section:additional_experiments_and_results}

In this section, we explore an additional modeling task, computational protein design, with its implementation details being discussed in Appendix \ref{section:implementation_details}.

\textbf{CPD, Node Classification.} Computational protein design (CPD) investigates a method's ability to design native-like protein sequences. In our CPD experiments, we explore fixed-backbone sequence design, where methods are provided with the 3D backbone structure of a protein and asked to generate a corresponding sequence. We train and evaluate each CPD method on the CATH 4.2 dataset created by \cite{ingraham2019generative}. This dataset contains 18,204, 608, and 1,120 training, validation, and test proteins, respectively, where all available protein structures with 40\% nonredundancy are partitioned by their CATH (class, architecture, topology/fold, homologous superfamily) classification. Baseline comparison methods for this task include a mixture of state-of-the-art Transformers, GNNs, and ENNs.

Under the assumption that native sequences are optimized for their structures \cite{kuhlman2000native}, the metrics with which we evaluate each method measure how well a method can distinguish a native-like sequence from a non-native sequence. In particular, following \cite{ingraham2019generative}, we adopt model perplexity as a measure of how well a method can model the language of native protein sequences. Similarly, we employ native sequence recovery (i.e., amino acid recovery) rates as a way of evaluating, on average, how well each method can design sequences that resemble native protein sequences.

\begin{table}[ht]
    \caption{Comparison of \textsc{GCPNet} with baseline methods for the CPD task. Results are reported in terms of the perplexity and amino acid recovery rates of each method for fixed-backbone sequence design. The best results for this task are in \textbf{bold}, and the second-best results are \underline{underlined}.}
    \label{table:cpd_results}
    \centering
    % \footnotesize  % Or footnotesize, scriptsize, tiny, etc.
    \begin{tabular}{lcccccc}
        & & & & & & \\ \midrule
        & & Perplexity\ $\downarrow$ & & & Recovery\ $\uparrow$ & \\
        \cmidrule(lr){2-4}\cmidrule(lr){5-7}
        Method & Short & Single & All & Short & Single & All \\ \midrule
        STran* \citeyearpar{ingraham2019generative} & 8.54 & 9.03 & 6.85 & 28.30 & 27.60 & 36.40 \\
        SGNN* \citeyearpar{jing2020learning} & 8.31 & 8.88 & 6.55 & 28.40 & 28.10 & 37.30 \\
        GVP* \citeyearpar{jing2021equivariant} & \underline{7.10} & \underline{7.44} & \underline{5.29} & 32.10 & 32.00 & 40.20 \\
        GBP* \citeyearpar{10.1145/3534678.3539441} & \textbf{6.14} & \textbf{6.46} & \textbf{5.03} & 33.22 & \textbf{33.22} & \textbf{42.70} \\ \midrule
        \textsc{GCPNet} w/o Frames & 7.71 & 8.18 & 5.87 & 31.82 & 31.72 & \underline{41.18} \\
        \textsc{GCPNet} w/o \textsc{ResGCP} & 9.63 & 9.88 & 7.09 & 27.27 & 27.02 & 35.27 \\
        \textsc{GCPNet} w/o Scalars & 18.51 & 18.37 & 18.17 & 8.70 & 8.55 & 8.62 \\
        \textsc{GCPNet} w/o Vectors & 10.41 & 10.53 & 8.87 & 26.42 & 26.23 & 28.99 \\
        \textsc{GCPNet} & 8.22 & 8.60 & 6.06 & \textbf{33.33} & \underline{32.86} & 40.32 \\ \midrule
    \end{tabular}
\end{table}

Table \ref{table:cpd_results} shows that, in representing proteins as amino acid residue-level graphs, \textsc{GCPNet} matches or exceeds the performance of several state-of-the-art prediction methods for CPD. In particular, \textsc{GCPNet} improves upon state-of-the-art short sequence recovery rates of previous methods by 0.5\% on average while maintaining competitive performance against other methods in all other metrics. We note that all CPD methods marked with * perform model inference autoregressively, introducing a significant computational bottleneck for real-world applications of these models. Inference with \textsc{GCPNet}, in contrast, is designed for direct prediction of amino acid sequences corresponding to a 3D protein structure, thereby decreasing inference runtime by more than a factor of two compared to other methods. While being a simple direct-shot prediction method for CPD, \textsc{GCPNet} is still able to achieve competitive results in terms of amino acid recovery rates for sequence generation, with reasonable results in terms of perplexity as well.

Interestingly, in the context of CPD, an ablation of our equivariant local frames $\bm{\mathcal{F}}_{ij}$ reveals that such frames are not useful for increasing \textsc{GCPNet}'s confidence in its structural understanding of the language of proteins (i.e., its perplexity). This suggests that future work could involve exploring alternative geometric encoding schemes for residue-based graphs when approaching the CPD task \cite{gao2022pifold} with \textsc{GCPNet}. This finding highlights the fact that the local frames $\bm{\mathcal{F}}_{ij}$ appear to be most useful in the context of representation learning on atomic graphs where lower-level molecular motifs are likely to appear, implying that future work towards improving CPD results with \textsc{GCPNet} could involve developing novel atom-level encoding schemes for residue-based graph predictions to leverage the promising results \textsc{GCPNet} yields in other dataset contexts. Nonetheless, our remaining ablations demonstrate that other design characteristics of \textsc{GCPNet} such as the $\mathbf{ResGCP}$ module and scalar and vector-valued feature representations enable \textsc{GCPNet} to better decode sequence-based information from 3D protein structures.

\begin{table}[ht]
\caption{Summary of \textsc{GCPNet}'s node and edge features for 3D input graphs derived for the LBA and PSR tasks. Here, $N$ and $E$ denote the number of nodes and edges in $\mathcal{G}$, respectively.}
\label{table:lba_psr_graph_features}
\centering
% \footnotesize  % Or footnotesize, scriptsize, tiny, etc.
\begin{tabular}{clll}
\toprule
                                             & \multicolumn{1}{c}{Feature} & \multicolumn{1}{c}{Type} & Shape \\ \midrule
\multirow{1}{*}{Node Features ($h$)}         & One-hot encoding of atom type         & Categorical (Scalar)               & $N \times 9$          \\
\multirow{1}{*}{Node Features ($\chi$)}      & Directional encoding of orientation   & Numeric (Vector)                   & $N \times 2$          \\ \midrule
\multirow{1}{*}{Edge Features ($e$)}         & Radial basis distance embedding       & Numeric (Scalar)                   & $E \times 16$         \\
\multirow{1}{*}{Edge Features ($\xi$)}       & Pairwise atom position displacement   & Numeric (Vector)                   & $E \times 1$          \\ \midrule
\multicolumn{1}{l}{\multirow{1}{*}{Total}}   & Node features                         &                                    & $N \times 11$         \\
\multicolumn{1}{l}{}                         & Edge features                         &                                    & $E \times 17$         \\ \bottomrule
\end{tabular}
\end{table}

\begin{table}[ht]
\caption{Summary of \textsc{GCPNet}'s node and edge features for 3D input graphs derived for the CPD task.}
\label{table:cpd_graph_features}
\centering
% \footnotesize  % Or footnotesize, scriptsize, tiny, etc.
\begin{tabular}{clll}
\toprule
                                             & \multicolumn{1}{c}{Feature} & \multicolumn{1}{c}{Type} & Shape \\ \midrule
\multirow{1}{*}{Node Features ($h$)}         & Dihedral angle encoding               & Numeric (Scalar)                   & $N \times 6$          \\
\multirow{1}{*}{Node Features ($\chi$)}      & Orientation and sidechain encoding    & Numeric (Vector)                   & $N \times 3$          \\ \midrule
\multirow{1}{*}{Edge Features ($e$)}         & Distance and positional embedding     & Numeric (Scalar)                   & $E \times 32$         \\
\multirow{1}{*}{Edge Features ($\xi$)}       & Pairwise atom position displacement   & Numeric (Vector)                   & $E \times 1$          \\ \midrule
\multicolumn{1}{l}{\multirow{1}{*}{Total}}   & Node features                         &                                    & $N \times 9$          \\
\multicolumn{1}{l}{}                         & Edge features                         &                                    & $E \times 33$         \\ \bottomrule
\end{tabular}
\end{table}

\begin{table}[ht]
\caption{Summary of \textsc{GCPNet}'s node and edge features for 3D input graphs derived for the NMS task.}
\label{table:nms_graph_features}
\centering
% \footnotesize  % Or footnotesize, scriptsize, tiny, etc.
\begin{tabular}{clll}
\toprule
                                             & \multicolumn{1}{c}{Feature} & \multicolumn{1}{c}{Type} & Shape \\ \midrule
\multirow{1}{*}{Node Features ($h$)}         & Invariant velocity encoding           & Numeric (Scalar)                   & $N \times 1$          \\
\multirow{1}{*}{Node Features ($\chi$)}      & Velocity and orientation encoding     & Numeric (Vector)                   & $N \times 3$          \\ \midrule
\multirow{1}{*}{Edge Features ($e$)}         & Edge and distance embedding           & Numeric (Scalar)                   & $E \times 17$         \\
\multirow{1}{*}{Edge Features ($\xi$)}       & Pairwise atom position displacement   & Numeric (Vector)                   & $E \times 1$          \\ \midrule
\multicolumn{1}{l}{\multirow{1}{*}{Total}}   & Node features                         &                                    & $N \times 4$          \\
\multicolumn{1}{l}{}                         & Edge features                         &                                    & $E \times 18$         \\ \bottomrule
\end{tabular}
\end{table}

\section{Implementation Details.}
\label{section:implementation_details}

\textbf{Featurization.}
\label{section:featurization}
As shown in Table \ref{table:lba_psr_graph_features}, for the LBA and PSR tasks, in each 3D input graph, we include as a scalar node feature an atom's type using a 9-dimensional one-hot encoding vector for each atom. As vector-valued node features, we include \textit{forward} and \textit{reverse} unit vectors in the direction of $x_{i + 1} - x_{i}$ and $x_{i - 1} - x_{i}$, respectively (i.e., the node's 3D orientation). For the input 3D graphs' scalar edge features, we encode the distance $\lVert x_{i} - x_{j} \rVert_{2}$ using Gaussian radial basis functions, where we use 16 radial basis functions with centers evenly distributed between 0 and 20 units (e.g., Angstrom). For the graphs' vector-valued edge features, we encode the unit vector in the direction of $x_{i} - x_{j}$ (i.e., pairwise atom position displacements).

As displayed in Table \ref{table:cpd_graph_features}, for the CPD task, in each 3D input graph, we include as scalar node features an encoding of each amino acid residue's dihedral angles $\{\sin, \cos\} \circ \{\phi, \psi, \omega\}$, where $\bm{\Phi}$, $\psi$, and $\omega$ are the dihedral angles computed from the corresponding protein's $C_{i - 1}$, $N_{i}$, $C_{i}$, and $N_{i + 1}$ backbone atoms. We then include as vector-valued node features each node's 3D orientation. For edge features, we use Gaussian radial basis function distance encodings as scalar edge features and pairwise atom position displacements as vector-valued edge features.

As illustrated in Table \ref{table:nms_graph_features}, for the NMS task, in each 3D input graph, we include as a scalar node feature an invariant encoding of each node's velocity vector, namely $\sqrt{v_{i}^{2}}$. Each node's velocity and orientation are encoded as vector-valued node features. Scalar edge features are represented as Gaussian radial basis distance encodings as well as the product of the charges in each node pair (i.e., $c_{i}c_{j}$). Lastly, vector-valued edge features are represented as pairwise atom position displacements.

\textbf{Hardware Used.}
\label{section:hardware_used}
The Oak Ridge Leadership Facility (OLCF) at the Oak Ridge National Laboratory (ORNL) is an open science computing facility that supports HPC research. The OLCF houses the Summit compute cluster. Summit, launched in 2018, delivers 8 times the computational performance of Titan’s 18,688 nodes, using only 4,608 nodes. Like Titan, Summit has a hybrid architecture, and each node contains multiple IBM POWER9 CPUs and NVIDIA Volta GPUs all connected with NVIDIA’s high-speed NVLink. Each node has over half a terabyte of coherent memory (high bandwidth memory + DDR4) addressable by all CPUs and GPUs plus 800GB of non-volatile RAM that can be used as a burst buffer or as extended memory. To provide a high rate of I/O throughput, the nodes are connected in a non-blocking fat-tree using a dual-rail Mellanox EDR InfiniBand interconnect. We used the Summit compute cluster to train all our models. For the LBA and NMS tasks, we used 16GB NVIDIA Tesla V100 GPUs for model training, whereas for the memory-intensive PSR and CPD tasks, we used 32GB V100 GPUs instead.

\textbf{Software Used.}
\label{section:software_used}
We used Python 3.8.12 \cite{10.5555/1593511}, PyTorch 1.10.2 \cite{NEURIPS2019_9015}, PyTorch Lightning 1.7.7 \cite{falcon2019pytorch}, and PyTorch Geometric 2.1.0post0 \cite{feypyg2019} to run our deep learning experiments. For each model trained, PyTorch Lightning was used to facilitate model checkpointing, metrics reporting, and distributed data parallelism across 6 V100 GPUs. A more in-depth description of the software environment used to train and run inference with our models is available at \href{https://github.com/BioinfoMachineLearning/GCPNet}{https://github.com/BioinfoMachineLearning/GCPNet}.

\textbf{Hyperparameters.}
\label{section:hyperparameters}
As shown in Tables \ref{table:lba_hyperparameter_search_space}, \ref{table:psr_hyperparameter_search_space}, \ref{table:cpd_hyperparameter_search_space}, and \ref{table:nms_hyperparameter_search_space}, we use a learning rate of $10^{-4}$ for all \textsc{GCPNet} models. The learning rate is kept constant throughout each model's training. For the NMS task, each model is trained for a minimum of 100 epochs and a maximum of 12,000 epochs. For all other tasks, each model is trained for a minimum of 100 epochs and a maximum of 1,000 epochs. For a given task, models with the best loss on the corresponding validation data split are then tested on the test split for the respective task. Note that, for the RS task, we do not perform any model hyperparameter tuning, following previous conventions from \cite{schneuing2023structurebased}.

\newpage

\begin{table}[ht]
    \caption{Hyperparameter search space for all \textsc{GCPNet} models through which we searched to obtain strong performance on the LBA task's validation split. The final parameters for the standard \textsc{GCPNet} model for the LBA task are in \textbf{bold}.}
    \label{table:lba_hyperparameter_search_space}
    \centering
    % \footnotesize  % Or footnotesize, scriptsize, tiny, etc.
    \begin{tabular}{lc}
        \toprule
        Hyperparameter                              &   Search Space \\ \midrule
        Number of \textsc{GCPNet} Layers            &   7, \textbf{8} \\
        Number of $\mathbf{GCP}$ Message-Passing Layers        &   \textbf{8} \\
        $\chi$ Hidden Dimensionality                &   \textbf{16}, 32 \\
        Learning Rate                               &   \textbf{0.0001}, 0.0003 \\
        Weight Decay Rate                           &   \textbf{0} \\
        $\mathbf{GCP}$ Dropout Rate                            &   \textbf{0.1}, 0.25 \\
        Dense Layer Dropout Rate                    &   \textbf{0.1}, 0.25 \\
        \bottomrule
    \end{tabular}
\end{table}

\begin{table}[ht]
    \caption{Hyperparameter search space for all \textsc{GCPNet} models through which we searched to obtain strong performance on the PSR task's validation split. The final parameters for the standard \textsc{GCPNet} model for the PSR task are in \textbf{bold}.}
    \label{table:psr_hyperparameter_search_space}
    \centering
    % \footnotesize  % Or footnotesize, scriptsize, tiny, etc.
    \begin{tabular}{lc}
        \toprule
        Hyperparameter                              &   Search Space \\ \midrule
        Number of \textsc{GCPNet} Layers            &   \textbf{5} \\
        Number of $\mathbf{GCP}$ Message-Passing Layers        &   \textbf{8} \\
        $\chi$ Hidden Dimensionality                &   \textbf{16}, 32 \\
        Learning Rate                               &   \textbf{0.0001}, 0.0003 \\
        Weight Decay Rate                           &   \textbf{0}, 0.0001 \\
        $\mathbf{GCP}$ Dropout Rate                            &   \textbf{0.1}, 0.25 \\
        Dense Layer Dropout Rate                    &   \textbf{0.1}, 0.25 \\
        \bottomrule
    \end{tabular}
\end{table}

\begin{table}[ht]
    \caption{Hyperparameter search space for all \textsc{GCPNet} models through which we searched to obtain strong performance on the CPD task's validation split. The final parameters for the standard \textsc{GCPNet} model for the CPD task are in \textbf{bold}.}
    \label{table:cpd_hyperparameter_search_space}
    \centering
    % \footnotesize  % Or footnotesize, scriptsize, tiny, etc.
    \begin{tabular}{lc}
        \toprule
        Hyperparameter                              &   Search Space \\ \midrule
        Number of \textsc{GCPNet} Encoder Layers    &   \textbf{9} \\
        Number of \textsc{GCPNet} Decoder Layers    &   \textbf{3} \\
        Number of $\mathbf{GCP}$ Message-Passing Layers        &   \textbf{8} \\
        $\chi$ Hidden Dimensionality                &   \textbf{16}, 32 \\
        Learning Rate                               &   \textbf{0.0001} \\
        Weight Decay Rate                           &   0.0, $\mathbf{10}^{\mathbf{-8}}$, 0.0001 \\
        $\mathbf{GCP}$ Dropout Rate                            &   0.1, \textbf{0.2}, 0.25, 0.4 \\
        Decoder Residual Updates                    &   False, \textbf{True} \\
        \bottomrule
    \end{tabular}
\end{table}

\begin{table}[ht]
    \caption{Hyperparameter search space for all \textsc{GCPNet} models through which we searched to obtain strong performance on the NMS task's validation split. The final parameters for the standard \textsc{GCPNet} model for the NMS task are in \textbf{bold}.}
    \label{table:nms_hyperparameter_search_space}
    \centering
    % \footnotesize  % Or footnotesize, scriptsize, tiny, etc.
    \begin{tabular}{lc}
        \toprule
        Hyperparameter                              &   Search Space \\ \midrule
        Number of \textsc{GCPNet} Layers            &   \textbf{4}, 7 \\
        Number of $\mathbf{GCP}$ Message-Passing Layers        &   \textbf{8} \\
        $\chi$ Hidden Dimensionality                &   \textbf{16} \\
        Learning Rate                               &   \textbf{0.0001}, 0.0003 \\
        Weight Decay Rate                           &   \textbf{0} \\
        $\mathbf{GCP}$ Dropout Rate                            &   0.0, \textbf{0.1} \\
        \bottomrule
    \end{tabular}
\end{table}




\end{appendices}

\end{document}